# Multiway *p*-spectral graph cuts on Grassmann manifolds

Dimosthenis Pasadakis[1] · Christie Louis Alappat[2] · Olaf Schenk[1] · Gerhard Wellein[2]

## Abstract

Nonlinear reformulations of the spectral clustering method have gained a lot of recent attention due to their increased numerical benefits and their solid mathematical background. We present a novel direct multiway spectral clustering algorithm in the *p*-norm, for $p \in (1, 2]$. The problem of computing multiple eigenvectors of the graph *p*-Laplacian, a nonlinear generalization of the standard graph Laplacian, is recast as an unconstrained minimization problem on a Grassmann manifold. The value of *p* is reduced in a pseudo-continuous manner, promoting sparser solution vectors that correspond to optimal graph cuts as *p* approaches one. Monitoring the monotonic decrease of the balanced graph cuts guarantees that we obtain the best available solution from the *p*-levels considered. We demonstrate the effectiveness and accuracy of our algorithm in various artificial test-cases. Our numerical examples and comparative results with various state-of-the-art clustering methods indicate that the proposed method obtains high quality clusters both in terms of balanced graph cut metrics and in terms of the accuracy of the labelling assignment. Furthermore, we conduct studies for the classification of facial images and handwritten characters to demonstrate the applicability in real-world datasets.

## 1 Introduction and related work

The act of creating clusters by segmenting a set into several parts is ever present in every scientific domain that deals with interacting or interconnected data. The formation of clusters consists of distributing a group of objects into distinct subsets. This process generally aims to obtain parts of roughly equal size with strong internal and weak external connections. Clustering using spectral methods is widely used, and can be applied to any kind of data with a suitable similarity metric between them forming a graphical structure. The theoretical background for this family of methods is based on the balanced graph partitioning problem, with reformulations connecting it also with random walks (Fountoulakis et al., 2019; Mahoney, 2012) and perturbation theory (Ng et al., 2001). As opposed to other popular clustering techniques, such as the k-means (MacQueen, 1967) and the expectation-maximization algorithm (Dempster et al., 1977), spectral methods perform well in nonconvex sample spaces, as they can avoid local minima (Bichot & Siarry, 2013). They have therefore been successfully applied in various fields of data clustering, such as computer vision (Malik et al., 2001), load balancing (Hendrickson & Leland, 1995), biological systems (Pentney & Meila, 2005) and text classification (Aggarwal & Zhai, 2012), and are a field of active research (Ge et al., 2021; Mizutani, 2021). Additionally, efficient variants employing multilevel techniques have been proposed (Dhillon et al., 2005, 2007). The authors refer to Jia et al. (2014) and Wierzchoń and Kłopotek (2018) for detailed overviews of various spectral graph clustering algorithms and recent advancements in the field.

Reformulating the spectral method from the traditional 2-norm to the $p$-norm has proven to lead to a sharp approximation of balanced cut metrics and improved clustering assignments (Amghibech, 2006; Gajewski & Gärtner, 2001). Such reformulations result in a tight relaxation of the spectral clustering problem, with the resulting solutions approximating closely the solution of the original discrete problem. The graph cut theoretically converges to the optimal Cheeger cut (Cheeger, 1969) for $p \to 1$, thus highlighting the superiority of $p$-spectral methods over their traditional 2-norm counterparts. These favorable properties have attracted a lot of recent attention. In Bühler and Hein (2009) partitions are obtained by thresholding the eigenvector associated with the second-smallest eigenvalue of the graph $p$-Laplacian, a nonlinear generalization of the graph Laplacian. In Jia et al. (2015), using the same objective function, a self-tuning $p$-spectral algorithm is proposed that determines the optimal value of $p$. The authors in Luo et al. (2010) generalize this approach to multiway partitioning by employing a modified gradient descent update that converges to multiple $p$-eigenvectors. The nodal properties of multiple eigenvectors of the graph $p$-Laplacian were investigated in Tudisco and Hein (2017). In Simpson et al. (2018) the authors introduce an explicit way to handle the constraints between the first two eigenvectors of the $p$-Laplacian, and propose a hybrid scheme to recursively partition large-scale graphs. Tight relaxations based on the concept of total variation, leading to similar sharp indicator eigenfunctions, have also been proposed for bi-partitioning (Szlam & Bresson, 2010; Bresson et al., 2012) and multiway problems (Bresson et al., 2013a; Hein & Setzer, 2011; Rangapuram et al., 2014). A monotonically descending adaptive algorithm for the minimization of total variation was proposed in Bresson et al. (2013b), and in Bresson et al. 2014), the concept of total variation was utilized in multiclass transductive learning problems. Reformulations of the spectral method in different $p$-norms have also been employed recently in local graph clustering methods (Fountoulakis et al., 2020; Liu &

Gleich, 2020), in hypergraph partitioning (Li et al., 2020), as well as in the clustering of signed graphs (Mercado et al., 2019).

The focus of our work is centered around developing a simple algorithm for direct multiway (or *k*-way) *p*-spectral clustering that effectively minimizes graph cuts. In doing so, we avoid problems that emerge from the greedy nature of recursive approaches and their lack of the global information of the graph (Simon & Teng, 1997). The existing direct multiway *p*-spectral approach for $p \in (1, 2)$ (Luo et al., 2010) relies on the computation of multiple eigenvectors of the graph *p*-Laplacian, which will subsequently be considered as the *p*-spectral coordinates of the nodes of the graph, by imposing mutual orthogonality between them. The constraint on the eigenvectors is enforced by means of a modified gradient descent minimization that might lead to approximation errors, as demonstrated later. This mutual orthogonality constraint, combined with the fact that the final partitioning step takes place in the reduced space of *p*-spectral coordinates, suggests that recasting the problem as an optimization procedure over a subspace that adheres by definition to these conditions, i.e., the Grassmann manifold, would be beneficial. The Riemannian structure of the Grassmann manifold and the development of robust optimization algorithms on it have been extensively researched (Edelman et al., 1999; Absil et al., 2007; Sato & Iwai, 2014).

### 1.1 Contributions and outline

This paper approaches spectral clustering from a different angle by reformulating it into a unconstrained minimization problem in the *p*-norm. We propose a new multiway *p*-spectral clustering method, and recast the problem of finding multiple eigenpairs of the graph *p*-Laplacian as a Riemannian optimization problem on a Grassmann manifold. From here on, we refer to the introduced algorithm as "pGrass". Our algorithm preserves the mutual orthogonality between the eigenvectors of the graph *p*-Laplacian (*p*-eigenvectors). We therefore succeed in reformulating a constrained minimization problem intro an unconstrained one on a manifold, a prevailing trend in optimization (Antoniou & Wu-Sheng, 2017). Special emphasis is put in the minimization of the nonlinear objective function which is achieved by means of a Grassmannian Newton method, with a truncated conjugate gradient algorithm for the linear intermediate steps (Huang et al., 2018). We reduce the value of *p* from $p = 2$ towards $p \approx 1$ in a pseudocontinuous manner that ensures that the majority of the evaluations take place close to $p \approx 1$, where the optimal results are expected to be found. Our algorithms is guaranteed to find the best available clustering solution in all the *p*-levels that are considered, by monitoring the monotonic reduction of the balanced graph cut metrics.

We provide five sets of numerical experiments in a total of 80 graphs, and compare our method against several state-of-the-art clustering algorithms. We begin with (i) a study of the effect the reduction of the value of *p* has on synthetic datasets. Here we also demonstrate the effectiveness of monitoring the monotonic reduction of our discrete graph cut objectives, and their correlation with the minimization of the gradient norm of our continuous objective. We proceed with (ii) synthetic tests that highlight the effect an increasing number of clusters has on the quality of the clustering. Next, for an artificial graph, we present (iii) the differences between the embeddings achieved by pGrass and standard spectral clustering methods in the 2-norm. Following the tests on artificial graphs, we conduct numerical experiments where we highlight the applicability of pGrass for real-world datasets. For the first experiment (iv) the classification of images containing facial expressions is presented. Finally, in (v), we apply pGrass in the problem of classifying handwritten characters from various languages according to their labels. Both real-world studies highlight the fact that depending on the way the *p*-eigenvectors are clustered in order to obtain the final discrete solution, our algorithm pGrass provides the best results either in terms of balanced graph cut metrics, or in terms of the accuracy of the labelling assignment.

In what follows, we recap the spectral graph clustering problem in Sect. 2. We initially define the metrics and matrices involved in spectral graph clustering. Then we outline the recursive bi-partitioning approach in the 2-norm and in the $p$-norm for $p \in (1, 2]$, and the traditional direct multiway spectral method. The process of solving an unconstrained minimization $p$-norm problem on a Riemannian manifold is presented in Sect. 3. Here we motivate our research for applying manifold minimization in order to find multiple eigenvectors of the graph $p$-Laplacian, we formulate our problem on the Grassmann manifold, and we present the key algorithmic components and optimization techniques of our algorithm. In Sect. 4 we present the performance of our algorithm in clustering artificial and real-world datasets with ground-truth labels, and, finally, in Sect. 5 we draw conclusions from this work and sketch future directions of research on the topic.

## 1.2 Notation

For the rest of this paper we denote scalar quantities with lower case, vectors by lower-case bold, sets by upper case, matrices with upper-case bold characters, and manifolds by upper-case calligraphic. The $p$-norm of a vector is defined as $\|\mathbf{u}\|_p$ with $p = 2$ being the Euclidean norm. The cardinality of a set $V$ is denoted by $|V|$, while for all other quantities $|\cdot|$ indicates their absolute value. The $i$th element of a vector $\mathbf{v}$ is denoted by $v_i$. The $i$th column vector of a matrix $\mathbf{V}$ is denoted by either $\mathbf{v}_i$, or $\mathbf{v}^i$. The latter is used in case that the subscript is occupied by the element index number or the norm of the vector. For example, when comparing the $i$th eigenvector computed in the 2 and the $p$-norm, we denote them as $\mathbf{v}_2^i$ and $\mathbf{v}_p^i$ respectively. The $(i, j)$th entry of a matrix $\mathbf{V}$ is symbolized by $v_{ij}$. The all-ones vector is denoted as $\mathbf{e}$, the identity matrix as $\mathbf{I}$, and the element-wise multiplication between matrices $\mathbf{A}$ and $\mathbf{B}$ as $\mathbf{A} \odot \mathbf{B}$.

## 2 Spectral graph clustering background

Graph clustering aims to distinguish groups of points according to their similarities. If these data points are defined by a matrix describing pointwise similarities, the problem of grouping them in $k$ parts is treated as a graph partitioning problem with an undirected weighted graph $\mathcal{G}(V, E, W)$ being constructed. Its nodes $V$ represent the data points, and the similarity between the connected edges $E$ is encoded in the elements $w_{ij} > 0$ of the weight matrix $\mathbf{W}$. Graph-theoretic approaches have proven to be highly successful in characterizing and extracting clusters. However, the resulting clustering problems frequently appear to be NP-hard (Wagner & Wagner, 1993).

Spectral clustering is a popular graph-based method due to the simplicity of its implementation, the reasonable computation time, and the fact that it overcomes the NP-hardness of other graph-theoretic approaches by solving a relaxed optimization problem in polynomial time. Its idea is based on the eigendecomposition of matrices that describe the connectivity of a graph (Bichot & Siarry, 2013). The spectral clustering of the total number of nodes $n = |V|$ into groups $C_1, \ldots, C_k$ is equivalent to a partitioning problem, usually with a dual objective: high intracluster similarity and low intercluster similarity is desired, while at the same time the vertex size $|C|$ (cardinality) or the volume $\text{vol}(C) = \sum_{i \in C} d_{ii}$ of the clusters should not differ excessively.

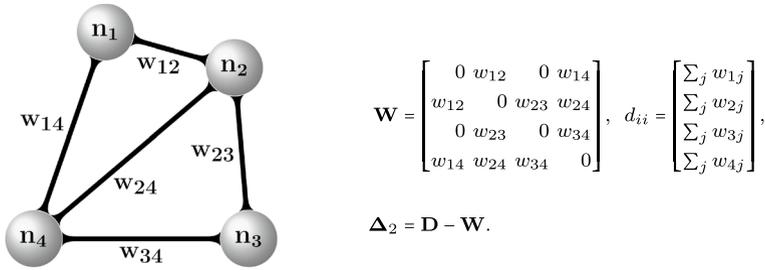

**Fig. 1** A simple, undirected, and connected graph $\mathcal{G}(V, E, W)$ with 4 vertices and 5 edges, with its adjacency **W**, degree **D**, and graph Laplacian $\mathbf{\Delta}_2$ matrices

## 2.1 Graphs, graph Laplacian, and graph cut metrics

The graph clustering objectives discussed previously are reflected in the balanced cut metrics presented below. When bisecting a graph $\mathcal{G}(V, E, W)$ into two subsets $C$ and its complement $\overline{C}$ $(= V\backslash C)$ the cut between them is defined as $\text{cut}(C, \overline{C}) = \sum_{i \in C, j \in \overline{C}} w_{ij}$. As balanced graph cut criteria we consider the ratio (Hagen & Kahng, 1991) and normalized cut (Shi & Malik, 2000), which in the case of bisection read

$$\text{RCut}(C, \overline{C}) = \frac{\text{cut}(C, \overline{C})}{|C|} + \frac{\text{cut}(\overline{C}, C)}{|\overline{C}|} \qquad (1)$$

$$\text{NCut}(C, \overline{C}) = \frac{\text{cut}(C, \overline{C})}{\text{vol}(C)} + \frac{\text{cut}(\overline{C}, C)}{\text{vol}(\overline{C})}. \qquad (2)$$

In the case of trying to identify $k$ clusters $C_1, \ldots, C_k$ in the entire node set $V$, the expressions are formalized as (Hagen & Kahng, 1992)

$$\text{RCut}(C_1, \ldots, C_k) = \sum_{i=1}^{k} \frac{\text{cut}(C_i, \overline{C_i})}{|C_i|} \qquad (3)$$

$$\text{NCut}(C_1, \ldots, C_k) = \sum_{i=1}^{k} \frac{\text{cut}(C_i, \overline{C_i})}{\text{vol}(C_i)}. \qquad (4)$$

The graph cut criteria discussed here describe nearly optimal clusters when their value approaches zero.

In spectral methods, the connectivity of $\mathcal{G}$ is usually described by means of the 2-norm graph Laplacian matrix $\mathbf{\Delta}_2 \in \mathbb{R}^{n \times n}$. The graph Laplacian matrix $\mathbf{\Delta}_2$ is a symmetric, positive semi-definite and diagonally dominant matrix whose spectral properties reveal a number of important topological properties of the graph (Bollobás, 1998; Chung, 1997). It is defined in terms of the adjacency matrix $\mathbf{W} \in \mathbb{R}^{n \times n}$ and the diagonal degree matrix $\mathbf{D} \in \mathbb{R}^{n \times n} \left( d_{ii} = \sum_{j=1}^{n} w_{ij} \right)$ as $\mathbf{\Delta}_2 = \mathbf{D} - \mathbf{W}$ (Fig. 1). Its normalized random walk counterpart is scaled by the degrees of the nodes and defined as $\Delta_2^{(n)} = \mathbf{D}^{-1}\Delta_2 = \mathbf{I} - \mathbf{D}^{-1}\mathbf{W}$.[1]

For a simple and undirected graph we additionally consider that $w_{ii} = 0$, and $w_{ij} = w_{ji}$. The graph Laplacian is often also realized as the linear operator whose action on a vector $\mathbf{u} \in \mathbb{R}^n$ induces the following quadratic form:

$$\langle \mathbf{u}, \Delta_2 \mathbf{u} \rangle = \mathbf{u}^T \Delta_2 \mathbf{u} = \frac{1}{2} \sum_{i,j=1}^n w_{ij}(u_i - u_j)^2, \tag{5}$$

demonstrating the positive semidefiniteness of $\Delta_2$. The eigenvalues of $\Delta_2$ can be ordered as $\lambda_1 \leq \lambda_2 \leq \cdots \leq \lambda_n$, with the eigenvector associated with $\lambda_1 = 0$ being the constant one, i.e., $\mathbf{v}^{(1)} = c \cdot \mathbf{e}$, where $c \in \mathbb{R}$.

In the following subsections we briefly describe three different spectral clustering methods. Section 2.2 discusses on the traditional spectral bi-partitioning approach, Sect. 2.3 on a $p$-norm extension of this bipartitioning method and finally Sect. 2.4 discusses on the multiway spectral clustering technique in the 2-norm.

### 2.2 Spectral bi-partitioning

In the case of bipartitioning, i.e., $k = 2$, we consider two complementary subsets $C, \overline{C}$ such that $C \cup \overline{C} = V$, $C \cap \overline{C} = \emptyset$. An indicator vector $\mathbf{u} = (u_1, \ldots, u_n)^T \in \mathbb{R}^n$ is defined for the vertex set $V = \{v_1, \ldots, v_n\}$ with

$$u_i = \begin{cases} \sqrt{\frac{|\overline{C}|}{|C|}} & \text{if } v_i \in C, \\ -\sqrt{\frac{|C|}{|\overline{C}|}} & \text{if } v_i \in \overline{C}. \end{cases} \tag{6}$$

The ratio cut partitioning metric (1) can now be expressed in terms of the graph Laplacian $\Delta_2$ with $\text{RCut}(C, \overline{C}) = \frac{\mathbf{u}^T \Delta_2 \mathbf{u}}{\mathbf{u}^T \mathbf{u}}$. Furthermore, it can be seen from (6) that the indicator vector of node assignments $\mathbf{u}$ is orthogonal to the constant vector $\mathbf{e}$, i.e., $\mathbf{u}^T \cdot \mathbf{e} = 0$. Therefore, the problem of minimizing the ratio cut (1) can be expressed as

$$\underset{C,\overline{C} \in V}{\text{minimize}} \frac{\mathbf{u}^T \Delta_2 \mathbf{u}}{\mathbf{u}^T \mathbf{u}}. \tag{7}$$

This optimization problem is $\mathcal{NP}$-hard, due to the discreteness of the indicator vector (6), thus a relaxation approach is followed by allowing $\mathbf{u}$ to attain values in all of $\mathbb{R}$, i.e., $u_i \in \mathbb{R}$. The relaxed optimization problem now reads

$$\underset{\mathbf{u} \in \mathbb{R}^n}{\text{minimize}} \frac{\mathbf{u}^T \Delta_2 \mathbf{u}}{\mathbf{u}^T \mathbf{u}} \tag{8a}$$

$$\text{subject to } \mathbf{u}^T \cdot \mathbf{e} = 0. \tag{8b}$$

---

[1] In what follows we refer to $\Delta_2^{(n)}$ as the normalized graph Laplacian. Note that it is different from the normalized symmetric graph Laplacian, as defined in Luxburg (2007).

The objective function (8a) is the Rayleigh quotient of the graph Laplacian matrix $\mathbf{\Delta}_2$. The minimum of the quotient is attained by the smallest eigenvalue $\lambda_1 = 0$ of $\mathbf{\Delta}_2$, with the associated eigenvector $\mathbf{v}^{(1)} = c \cdot \mathbf{e}$ being the minimizer. However, this eigenpair corresponds to the trivial partition $V = V \cup \emptyset$. Additionally, for nonconnected graphs, the multiplicity of $\lambda_1$ corresponds to the number of connected components. Therefore, taking into account the constraint (8b) we seek the second-smallest eigenvalue, called the algebraic connectivity of the graph (Fiedler, 1973), and its associated eigenvector. For a connected graph $\mathcal{G}$, this corresponds to $\mathbf{v}^{(2)}$, also termed Fiedler's eigenvector. It enables the partitioning of $\mathcal{G}$ into the two complementary sets $C, \overline{C}$ by thresholding its entries around zero, or their median value for tightly balanced partitioning applications. For a detailed analysis of the properties of the eigenspectrum of $\mathbf{\Delta}_2$ we refer the reader to Chung (1997). A computationally more expensive alternative, used more widely in clustering applications, is to perform a sweep cut on the Fiedler eigenvector by considering each of the $n$ cuts possible from the entries of $\mathbf{v}^{(2)}$ and selecting the one that minimizes the RCut (1). This process can be easily generalized to the normalized case $\mathbf{\Delta}_2^{(n)}$ (Luxburg, 2007), corresponding to a a minimization of the NCut (2).

Obtaining $k$-clusters from the spectral graph bisection method is possible by recursively bipartitioning the graph until the desired number of $k$ clusters is reached. At each recursive step, the partition whose bisection leads to smaller values of the global ratio cut (3) is split into two. Alternatively, in order to directly realize multiple strongly connected components of $\mathcal{G}$ the procedure outlined Sect. 2.4 is followed.

### 2.3 Bi-partitioning with the graph *p*-Laplacian

Reformulating spectral graph partitioning in the *p*-norm, for $p \in (1, 2]$, is based on the fact that better theoretical bounds on the balanced partitioning metrics, introduced in Sect. 2.1, are achieved at the limit $p \to 1$. A $p$ approaches one, the resulting bi-partition indicator vector $\mathbf{u}$ attains more discrete values and leads to nearly optimal balanced graph cut metrics and tighter partitionings (Gajewski & Gärtner, 2001; Amghibech, 2003; Bühler & Hein, 2009). At the limit of $p = 1$, solving the total variation problem (Szlam & Bresson, 2010; Bresson et al., 2012) has also been proven to be a tighter relaxation for balanced discrete graph cut metrics than the 2-norm relaxation (7).

The graph Laplacian operator $\mathbf{\Delta}_2$ can be redefined in the *p*-norm. For a node $i \in V$ the *p*-Laplacian operator $\mathbf{\Delta}_p$ is defined as $(\mathbf{\Delta}_p \mathbf{u})_i = \sum_{j \in V} w_{ij} \phi_p (u_i - u_j)$ and its normalized counterpart as $\left(\mathbf{\Delta}_p^{(n)} \mathbf{u}\right)_i = \frac{1}{d_i} \sum_{j \in V} w_{ij} \phi_p (u_i - u_j)$, with $\phi_p : \mathbb{R} \to \mathbb{R}$ being $\phi_p(x) = |x|^{p-1} \text{sign}(x)$, for $x \in \mathbb{R}$. In what follows we focus on the standard graph *p*-Laplacian case, but all concepts can be easily generalized to the normalized case. The *p*-Laplacian operator is nonlinear, with $\mathbf{\Delta}_p(\gamma \mathbf{x}) \neq \gamma \mathbf{\Delta}_p(\mathbf{x})$ for $\gamma \in \mathbb{R}$ and $p \in (1, 2)$, and the linear counterpart $\mathbf{\Delta}_2$ is recovered for $p = 2$, as $\phi_2(x) = x$ and $\mathbf{\Delta}_2(\cdot) = \mathbf{\Delta}_p(\cdot)$. Therefore the action of the standard graph Laplacian operator on a vector $\mathbf{u} \in \mathbb{R}^n$ can be generalized in the *p*-norm as $\langle \mathbf{u}, \mathbf{\Delta}_p \mathbf{u} \rangle = \frac{1}{2} \sum_{i,j=1}^{n} w_{ij} |u_i - u_j|^p$.

Similar to the approach followed in Sect. 2.2, we wish to obtain the second-smallest eigenvector of the symmetric graph *p*-Laplacian $\mathbf{\Delta}_p \in \mathbb{R}^{n \times n}$ in order to minimize the value of the RCut (1). The Rayleigh-Ritz principle, extended to the nonlinear case, states that a scalar value $\lambda_p \in \mathbb{R}$ is called an eigenvalue of $\mathbf{\Delta}_p$ if there exists a vector solution $\mathbf{v} \in \mathbb{R}^n$ such that $(\mathbf{\Delta}_p \mathbf{v})_i = \lambda_p \phi_p(v_i)$ with $i = 1, \ldots, n$. In order to obtain the smallest eigenpair of the $p$-Laplacian operator, we reformulate the Rayleigh quotient minimization problem from the linear 2-norm case $F_2(\mathbf{u}) : \mathbb{R}^n \to \mathbb{R}$,

$$F_2(\mathbf{u}) = \frac{\langle \mathbf{u}, \mathbf{\Delta}_2 \mathbf{u} \rangle}{\|\mathbf{u}\|_2^2} = \frac{1}{2} \frac{\sum_{i,j=1}^n w_{ij}(u_i - u_j)^2}{\|\mathbf{u}\|_2^2}, \tag{9}$$

to the nonlinear one $F_p(\mathbf{u}) : \mathbb{R}^n \to \mathbb{R}$ as

$$F_p(\mathbf{u}) = \frac{\langle \mathbf{u}, \mathbf{\Delta}_p \mathbf{u} \rangle}{\|\mathbf{u}\|_p^p} = \frac{1}{2} \frac{\sum_{i,j=1}^n w_{ij}|u_i - u_j|^p}{\|\mathbf{u}\|_p^p} \tag{10}$$

with the $p$-norm defined as $\|\mathbf{u}\|_p = \sqrt[p]{\sum_{i=1}^n |u_i|^p}$. A vector $\mathbf{v} \in \mathbb{R}^n$ is an eigenvector of $\mathbf{\Delta}_p$ if and only if it is a critical point of (10) (Bhatia, 1997). The associated $p$-eigenvalue is given by $F_p(\mathbf{v}) = \lambda_p$. The functional $F_p$ is nonconvex, and it is easy to notice that for some scalar $\gamma \in \mathbb{R}$ it is invariant under scaling, and thus $F_p(\gamma \mathbf{u}) = F_p(\mathbf{u})$.

Additionally, fundamental properties of the graph Laplacian in the linear case $p = 2$, which relate the eigenspectrum of $\mathbf{\Delta}_2$ to the algebraic connectivity of the graph (Fiedler, 1973), can be extended to the nonlinear one with $p \in (1, 2]$. The multiplicity of the smallest $p$-eigenvalue $\lambda_p^{(1)} = 0$ corresponds to the number of connected components in the graph (Bühler & Hein, 2009), and the associated eigenvector is constant. Therefore, for a connected graph, we are searching for the second eigenvalue $\lambda_p^{(2)}$ of $F_p$ and the associated eigenvector $\mathbf{v}^{(2)}$ in order to obtain a bi-partition. Furthermore, any two eigenvectors $\mathbf{v}^{(\alpha)}, \mathbf{v}^{(\beta)}$, with $\alpha \neq \beta$, of the $p$-Laplacian operator associated with nonzero eigenvalues are approximately $p$-orthogonal (Luo et al., 2010), i.e., $\sum_i \phi_p(v_i^{(\alpha)}) \phi_p(v_i^{(\beta)}) \approx 0$.

### 2.4 Direct multiway spectral clustering

Exploiting information from $k$ eigenvectors of the graph Laplacian matrix $\mathbf{\Delta}_2$ allows the direct $k$-way partitioning of a graph into $C_1, \ldots, C_k$ clusters, thus circumventing the need for a recursive strategy.

A relaxation approach is followed again for the minimization of RCut (3). We define $k$ indicator vectors $\mathbf{u}_j = (u_{1,j}, \ldots, u_{n,j})^T$ such that for $i = \{1, \ldots, n\}, j = \{1, \ldots, k\}$,

$$u_{i,j} = \begin{cases} \frac{1}{\sqrt{|C_j|}} & \text{if } v_i \in C_j, \\ 0 & \text{otherwise.} \end{cases} \tag{11}$$

The matrix $\mathbf{U} \in \mathbb{R}^{n \times k}$ contains these $k$ orthonormal vectors in its columns, thus $\mathbf{U}^T \mathbf{U} = \mathbf{I}$. The expression for estimating the global ratio cut (3) is now

$\text{RCut}(C_1, \ldots, C_k) = \text{Tr}(\mathbf{U}^T \mathbf{\Delta}_2 \mathbf{U})$ with Tr being the trace of a matrix. The discrete optimization problem for the minimization of (3) reads

$$\underset{C_1, \ldots, C_k}{\text{minimize}} \ \text{Tr}(\mathbf{U}^T \mathbf{\Delta}_2 \mathbf{U}). \tag{12a}$$

Finding globally optimum solutions for this expression is again a known NP-hard problem. The optimization problem is therefore relaxed by allowing the entries of matrix $\mathbf{U}$ to attain any value in $\mathbb{R}$, i.e., $\mathbf{u}_j \in \mathbb{R}^n$. The relaxed optimization problem now reads

$$\underset{\mathbf{U}\in\mathbb{R}^{n\times k}}{\text{minimize}}\ F_2(\mathbf{U}) = \text{Tr}\left(\mathbf{U}^\text{T}\mathbf{\Delta}_2\mathbf{U}\right) \tag{13a}$$

$$\text{subject to } \mathbf{U}^\text{T}\mathbf{U} = \mathbf{I}. \tag{13b}$$

Fan's trace min/max principle (Bhatia, 1997) dictates that the solution to this minimization problem is given by a matrix $\mathbf{U}$ whose first $k$ columns are spanned by the eigenvectors associated with the $k$ smallest eigenvalues of $\mathbf{\Delta}_2$. In order to obtain discrete clusters from the resulting real valued eigenvectors we consider for the $n$ nodes of the graph $n$ vectors $\mathbf{h}_i = \mathbf{U}_i^\text{T} \in \mathbb{R}^k\ \forall i \in [1,n]$. These are considered the spectral coordinates of the graph and have to be divided into $k$-groups $C_1, \ldots, C_k$. Similar to the bisection approach, this procedure can be generalized to the normalized functional $F_2^{(n)}$ that minimizes the NCut (4). In this case the indicator vectors are defined as $u_{i,j} = \begin{cases} \frac{1}{\sqrt{\text{vol}(C_j)}} & \text{if } v_i \in C_j, \\ 0 & \text{otherwise.} \end{cases}$ With the relaxed constraint now reading $\mathbf{U}^\text{T}\mathbf{D}\mathbf{U} = \mathbf{I}$, the $k$ columns of the matrix solution $\mathbf{U}$ correspond to the eigenvectors spanned by the $k$ smallest eigenvalues of $\mathbf{\Delta}_2^{(n)}$. This is the normalized spectral clustering approach introduced in Shi and Malik (2000).

As a final remark here, we note that the functional $F_2$ is invariant under a change of basis, i.e., $F_2(\mathbf{U}\mathbf{Q}) = F_2(\mathbf{U})$, for all $\mathbf{Q}$ belonging to the group of $k \times k$ orthogonal matrices, $\mathcal{O}(k) = \{\mathbf{Q} \in \mathbb{R}^{k\times k}\ |\ \mathbf{Q}^\text{T}\mathbf{Q} = \mathbf{I}\}$. This property will enable the reformulation of our $p$-spectral clustering problem into an unconstrained manifold minimization problem in Sect. 3.

## 3 A Grassmannian optimization approach to *p*-spectral clustering

The following section is devoted to the introduction of our multiway *p*-spectral clustering algorithm on Grassmann manifolds. In Sect. 3.1 we present some practical evidence that motivate the development of our method. In Sect. 3.2 we introduce the formulation of the unconstrained minimization problem that leads to the clustering of graphs. In Sect. 3.3 we discuss the optimization techniques employed in the algorithm, and finally in Sect. 3.4 we present the different ways the final discretization of the *p*-eigenvectors is achieved. The entire algorithmic scheme is summarized in Sect. 3.5.

### 3.1 Motivation

Besides the theoretical advantages of performing spectral bipartitioning in the *p*-norm, briefly discussed in Sect. 2.3, we further show a practical consideration that motivates our research on *p*-spectral clustering. In order to demonstrate this we calculate the second eigenvector of the graph Laplacian $\mathbf{\Delta}_2$ and the graph *p*-Laplacian $\mathbf{\Delta}_p$ for the graph of the 2-dimensional (2D) finite element mesh "grid1_dual" from the AG-Monien Graph Collection (Diekmann & Preis, 2018), with 224 nodes and 420 edges, and attempt to extract two clusters ($k = 2$) from the entries of the second eigenvector by thresholding it around zero. The results are illustrated in Fig. 2. We plot the mesh (graph) on the horizontal axis (*x*, *y* coordinates) and the eigenvector entries on the vertical one (*z* coordinate). Each eigenvector entry is visualized using the *x* and *y* coordinates of the associated node of the mesh in order to demonstrate the correspondence between graph Laplacian eigenvectors and graph clusters. In the standard spectral computations ($p = 2$) the entries of the Fiedler eigenvector $\mathbf{v}^{(2)}$ are distributed uniformly around zero. The number of cut edges is 20 and the value of the RCut = 0.179. In contrast, the entries of the second $p$-eigenvector $\mathbf{v}_p^{(2)}$ for $p = 1.1$ are organized into two easily distinguishable partitions, while at the same time the size of the edge cut is reduced to 16, and the value of the RCut = 0.143. The reason for this improved performance in the $p$-norm is the fact that as $p \to 1$, the cut obtained by thresholding $\mathbf{v}_p^{(2)}$ approaches its optimal value (Amghibech, 2006; Bühler & Hein, 2009). When considering multiple eigenvectors ($k > 2$) this tendency towards optimal cut values as $p$ approaches one has been proven for graphs for which the number of strong nodal domains of the eigenvector corresponding to the $k$-th smallest eigenvalue $\lambda_k$ is equal to $k$ (Tudisco & Hein, 2017), e.g., the unweighted path graph. However, the application of $p$-Laplacian direct multiway clustering in more general graphs has shown promising results (Luo et al., 2010).

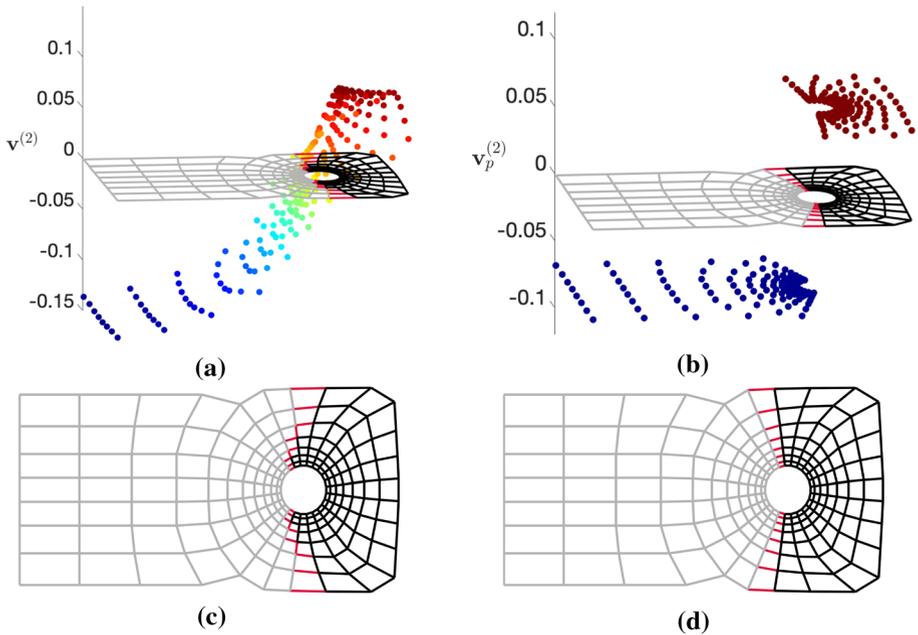

**Fig. 2** Finding two clusters based on the entries of the second eigenvector of the graph Laplacian $\Delta_2$ and of the graph $p$-Laplacian $\Delta_p$ for a finite element mesh (see text for details). The two partitions are depicted in black and gray, while the cut edges are depicted in red. The $z$-axis represents the value of the entries of the eigenvector, with their coloring indicating their distance from zero. **a** Standard spectral computation ($p = 2$). **b** Spectral computation in the $p$-norm for $p = 1.1$. **c** The standard spectral clusters. **d** The $p$-spectral clusters for $p = 1.1$. (Best viewed in color.)

In order to motivate the computation of multiple $p$-eigenvectors we additionally consider the fact that recursive bisection is highly dependant on the decisions made during the early stages of the process. Additionally, recursive methods suffer from the lack of global information as they do not optimize over the entire node set in order to find $k$ optimal partitions, but instead focus on finding optimal bisections at each recursive step. Thus, they may result in suboptimal partitions (Simon & Teng, 1997). This necessitates the further advancement of methods for direct multiway $p$-spectral clustering.

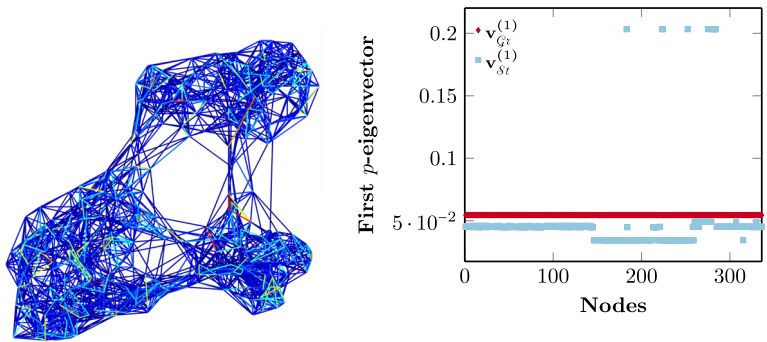

**Fig. 3** Values of the entries of the first eigenvector $\mathbf{v}^{(1)}$ of $\Delta_p$ for the *E. coli* graph (illustrated), after minimizing the functional (14a) over the Stiefel $\mathcal{St}$ and the Grassmann $\mathcal{Gr}$ manifold. The graph in question is connected and thus $\mathbf{v}^{(1)}$ should be constant. This behavior is observed only on $\mathcal{Gr}$ (in red), as $\mathbf{v}^{(1)}$ does not converge to a constant vector on $\mathcal{St}$ (in blue)

### 3.2 Direct multiway *p*-spectral clustering

Taking into account the objective function for spectral bipartitioning in the *p*-norm (10), the relaxed optimization problem of estimating multiple eigenvectors of the graph Laplacian (13) can be reformulated in the *p*-norm as

$$\underset{\mathbf{U}\in\mathbb{R}^{n\times k}}{\text{minimize}}\ F_p(\mathbf{U}) = \sum_{l=1}^{k}\sum_{i,j=1}^{n}\frac{w_{ij}|u_i^l - u_j^l|^p}{2\|\mathbf{u}^l\|_p^p} \quad (14a)$$

$$\text{subject to}\ \sum_{i=1}^{n}\phi_p(u_i^l)\phi_p(u_i^m) = 0\ \ \forall l\neq m,\ p\in(1,2],\ l\in[1,k],\ m\in[1,k].\quad (14b)$$

The cluster indices are denoted by $l, m = 1, 2, \ldots, k$. The final number of clusters $k$ is considered predetermined in this work. The matrix $\mathbf{U} = (\mathbf{u}_1, \ldots, \mathbf{u}_k)$ contains the eigenvectors associated with the smallest $k$ eigenvalues of the *p*-Laplacian operator $\Delta_p$ in its columns. In the case of normalized *p*-spectral clustering the normalized functional reads $F_p^{(n)}(\mathbf{U}) = \sum_{l=1}^{k}\sum_{i,j=1}^{n}\left(w_{ij}|u_i^l - u_j^l|^p\right)/\left(2d_i\|\mathbf{u}^l\|_p^p\right)$. This scaling by the degree $d_i$ of the corresponding row $i$ results in the matrix $\mathbf{U}$ containing the eigenvectors of the normalized *p*-Laplacian operator $\Delta_p^{(n)}$ (see Sect. 2.3) in its columns. For brevity we restrict our analysis in this section in the case of unnormalized *p*-spectral clustering.

The constraint for *p*-orthogonal eigenvectors (14b) renders the optimization problem intractable. Therefore, we replace it with the traditional constraint $\mathbf{U}^\mathsf{T}\mathbf{U} = \mathbf{I}$ (13b), a tight approximation as shown in Luo et al. (2010). This constraint corresponds to the Stiefel manifold, which is composed of all orthogonal column matrices

$$\mathcal{St}(k, n) = \{\mathbf{U} \in \mathbb{R}^{n\times k} \mid \mathbf{U}^\mathsf{T}\mathbf{U} = \mathbf{I}\}. \quad (15)$$

That is, a point in the Stiefel manifold is a specific orthogonal matrix (Edelman et al., 1999). Similar to standard direct *k*-way spectral clustering, we are interested in converging to some orthonormal basis of the eigenspace and not on the exact eigenvectors (Luxburg, 2007). The final transformation of the *p*-spectral coordinates into clusters is performed by either a flat algorithm like k-means or by rotating the normalized eigenvectors as shown in Yu and Shi (2003). Both algorithms are based on the relative distances between points and not on the exact values of their coordinates. However, every set of orthonormal eigenvectors forming the matrix **U** is considered to be unique on $\mathcal{S}t$, even if they correspond to the same basis. Therefore, optimizing our objective (14a) over the Stiefel manifold leads to the well known identifiability issue (Wang et al., 2017), with the redundantly big search space of the Stiefel manifold causing slow convergence and the increased probability of getting stuck in local minima for a nonconvex function. This behavior is illustrated in Fig. 3 for the problem of finding the first constant eigenvector of the graph $p$-Laplacian $\Delta_p$ for a graph representing the Ecoli dataset from the UCI dataset collection (Dua & Graff, 2017). Optimizing over the Stiefel manifold (in cyan) leads to a failure to converge to the known constant solution $\mathbf{v}_p^{(1)} = c\mathbf{1}$ (Bühler & Hein, 2009). Thus, in this case, additional constraints have to be imposed, i.e., the Stiefel gradient corresponding to the first eigenvector has to be set to zero in order for it to attain constant values. This gradient correction approach guides the algorithm towards the correct solution, but is not applicable to the rest of the $k-1$ eigenvectors of $\Delta_p$ as there is no theoretical guarantee for the values they should attain.

We thus consider the group of all $k \times k$ orthogonal matrices $\mathcal{O} = \{Q \in \mathbb{R}^{k \times k} \mid \mathbf{Q}^\mathsf{T}\mathbf{Q} = \mathbf{I}\}$. Searching for $k$ nonspecific and mutually orthogonal vectors as the solution to (14a) means that two solutions $\mathbf{U}_1$ and $\mathbf{U}_2$ belonging to the Stiefel manifold are considered equivalent if there exists some $\mathbf{Q} \in \mathcal{O}(k)$ such that $\mathbf{U}_1 = \mathbf{U}_2\mathbf{Q}$. This corresponds to the Grassmann manifold, a quotient space of $\mathcal{S}t(k, n)$ (Sato & Iwai, 2014), defined as

$$\mathcal{G}r(k, n) \simeq \mathcal{S}t(k, n)/\mathcal{O}(k) = \{\mathrm{span}(\mathbf{U}) : \mathbf{U} \in \mathbb{R}^{n \times k}, \mathbf{U}^\mathsf{T}\mathbf{U} = \mathbf{I}\}. \tag{16}$$

Points on $\mathcal{G}r(k, n)$ are understood as linear subspaces represented by an arbitrary basis stored as an $n$-by-$k$ orthonormal matrix (Edelman et al., 1999). The choice of the matrix **U** for these points is not unique, unlike for the ones on $\mathcal{S}t(k, n)$, with points on $\mathcal{G}r$ being defined through the relationship

$$\mathbf{U}^{\mathcal{G}r} = \{\mathbf{U}\mathbf{Q} \mid \forall\, \mathbf{Q} \in \mathcal{O}(k)\}, \quad \mathbf{U} \in \mathbb{R}^{n \times k},\, n \gg k. \tag{17}$$

Optimizing our objective over the Grassmann manifold results in a reduced search space, with the solutions being an approximation of the orthonormal eigenvectors of $\Delta_p$, satisfying fundamental properties of spectral graph theory, as outlined in Sect. 2.3, without imposing additional constraints. This behavior can be observed in Fig. 3, where optimizing over the Grassmann manifold leads to the constant first eigenvector of $\Delta_p$ (in red) for the Ecoli dataset.

Thus, we approximate function (14a) as being invariant to any choice of basis and only depending on the subspace spanned by the $p$-eigenvectors, i.e., the columns of **U**. The optimization problem of (14a) can now be reformulated as an unconstrained problem on the Grassmann manifold as follows:

$$\operatorname*{minimize}_{\mathbf{U}\in\mathcal{G}r(k,n)} F_p(\mathbf{U}) = \sum_l^k \sum_{i,j=1}^n \frac{w_{ij}|u_i^l - u_j^l|^p}{2\|\mathbf{u}^l\|_p^p}, \quad p \in (1,2]. \tag{18}$$

### 3.3 Optimization techniques

Section 3.2 revealed that the direct multiway $p$-spectral clustering problem can be approximated as an optimization problem on a Grassmann manifold. Manifold optimization has been extensively developed over the last couple of decades, with the intention of providing robust numerical algorithms for problems on subspaces with a Riemannian structure. The work of Absil et al. (2007) and Edelman et al. (1999) set the foundation to analyze such problems, with a focus on establishing a theory that leads to efficient numerical algorithms on the Stiefel $\mathcal{S}t(k,n)$ and Grassmann $\mathcal{G}r(k,n)$ manifolds. Specifically, they determine the Riemannian gradient and Hessian as the most critical ingredients in order to design first and second order algorithms on these subspaces. In particular, the Riemannian gradient and Hessian are projections of their Euclidean counterparts onto the tangent space of the manifold and the mapping between them is well established. Thus, in our case, the primary inputs to the manifold optimisation routines are the functional $F_p$ (18) along with its Euclidean gradient and optionally Hessian when using second order algorithms.

The entries of the Euclidean gradient ($\mathbf{g}^k$) of $F_p$ with respect to $u_m^k$ read[2]

$$g_m^k = \frac{\partial F_p}{\partial u_m^k} = \frac{p}{\|\mathbf{u}^k\|_p^p}\left[\sum_{j=1}^n w_{mj}\phi_p\left(u_m^k - u_j^k\right) - \phi_p(u_m^k)\sum_{i,j=1}^n \frac{w_{ij}|u_i^k - u_j^k|^p}{2\|\mathbf{u}^k\|_p^p}\right]. \tag{19}$$

The Hessian of the functional is not sparse and can cause storage and scaling problems for big problem sizes. Hence, we use a sparse approximation of the Hessian by discarding the low rank terms as shown in Bühler and Hein (2009). The Euclidean Hessian follows the sparsity pattern of $\mathbf{W}$ and is approximated as

$$h_{ml}^k = \frac{\partial g_m^k}{\partial u_l^k} \approx \begin{cases} \frac{p(p-1)}{\|\mathbf{u}^k\|_p^p}\sum_{j=1}^n w_{mj}|u_m^k - u_j^k|^{p-2} & \text{if } m=l, \\ \frac{-p(p-1)}{\|\mathbf{u}^k\|_p^p}w_{ml}|u_m^k - u_l^k|^{p-2} & \text{otherwise.} \end{cases} \tag{20}$$

Our objective function $F_p(\mathbf{U})$ (18) is nonconvex for $p \in (1,2)$, and thus convergence to a global minimum cannot be guaranteed. Minimizing $F_p$ directly for a small value of $p$ results, in most cases, in convergence to a nonoptimal local minimum. Therefore, we take advantage of the fact that our minimization problem (18) exhibits a convex behavior for $p = 2$, and thus the global minimizer can be computed. The fact that $F_p$ is also continuous in $p$ suggests that for close values of $p_1, p_2$, the solution of $F_{p_1}(\mathbf{U}), F_{p_2}(\mathbf{U})$ will be close as well (Bühler & Hein, 2009). Accordingly, to find a solution at a given $p \in (1, 2)$ we solve (18) by gradually reducing the value of $p$ (starting from $p = 2$), with the solution at the current $p$ serving as the initial iterate for the next $p$-level. In previous works (Bühler & Hein, 2009; Luo et al., 2010) the value of $p$ was decreased linearly. We instead decrease $p$ in a pseudocontinuous fashion, inspired by second order interior point methods and the way they handle the barrier parameter in order to achieve a superlinear rate of convergence (Byrd et al., 1998). The update rule for the value of $p$ reads:

$$p = 1 + \max\left(\text{tol}, \min\left(\kappa \cdot (p-1), (p-1)^\theta\right)\right), \tag{21}$$

with $\kappa \in (0, 1), \theta \in (1, 2)$, and tol $= 10^{-1}$. The lower bound of this update rule is $p \geq 1 + \text{tol}$, thus avoiding numerical instabilities with the discontinuity at $p = 1$. The value of $p$ is decreased at a superlinear rate, with the majority of the evaluations taking place close to $p = 1$, where the highest quality clusters are expected to be obtained.

---

[2] See "Appendix A" for the detailed derivation.

In each level of $p$, we minimize our objective with a Grassmannian Newton's method, as it has proven to have a superlinear convergence rate close to the local optima and quadratic elsewhere (Absil et al., 2007). The linear substeps within the Newton method are handled by a Grassmannian truncated conjugate gradient scheme (Antoniou & Wu-Sheng, 2017). For sparse or banded adjacency matrices $\mathbf{W}$ with bandwidth $2q + 1$ the computational cost per Newton iteration on the Grassmann is $O(nq^2k) + O(nk^2)$, assuming that $k, q \ll n$. Such matrices are commonly encountered in practical real-world applications. If the bandwidth is significantly narrow, i.e. $q^2 \approx k$, or if $\mathbf{W}$ is tridiagonal then the cost becomes $O(nk^2)$ (Absil et al., 2004). This reduction in the cost per iteration suggests that for very large and sparse adjacency matrices one can exploit the benefits from reordering methods that reduce the bandwidth size (Davis, 2006).

We use the Riemannian optimization software package ROPTLIB (Huang et al., 2018) to perform the Grassmannian Newton's steps. The Newton's minimization procedure is terminated if the norm of the gradient ($\|\mathbf{g}_m^k\|$) at iteration $m$ is close to zero, i.e., $\|\mathbf{g}_m^k\|/\|\mathbf{g}_0^k\| < 10^{-6}$. In addition to the stopping criteria for Newton's method within each $p$ level we also use an additional global stopping criterion based on cut values (RCut or NCut) at each $p$ level. If the cut value increases by at least 5% compared to its value at the previous $p$ level we terminate the algorithm and choose the cluster corresponding to the smallest cut value, thus ensuring the semi-monotonic descent of our discrete objective.

### 3.4 Discretizing the *p*-eigenvectors

Similar to multiway spectral clustering in 2-norm (see Sect. 2.4), the final clustering solution is obtained by discretizing the multiple *p*-eigenvectors, stored in the matrix **U**, obtained by solving the Grassmannian optimization problem (18). Various approaches have been proposed for this discretization in the 2-norm (Wierzchoń & Kłopotek, 2018; Verma & Meila, 2005).

We consider two different methods for the discretization of the *p*-eigenvectors. The first one is k-means, which is the most commonly used algorithm for the clustering of eigenvectors. However, the results of k-means depend heavily on the initial guess, and, therefore, in general k-means is run multiple times with different initial guesses and the best result is picked. We follow the approach of Verma and Meila (2005) with multiple orthogonal and random initial guesses that generally lead to a stable result. The second algorithm that we employ for the clustering of the *p*-eigenvectors is applicable only when minimizing the NCut graph cut metric. It is based on the fact that the application of a rotation matrix **P** transforms the matrix **U**, containing the normalized *p*-eigenvectors in its columns, into a cluster indicator matrix containing only one nonzero entry per row that indicates the cluster index (Wierzchoń & Kłopotek, 2018). We consider the set of indicator matrices $J = \{\mathbf{J} \in \{0,1\}^{n \times k} : \mathbf{J} \cdot \mathbf{e}_k = \mathbf{e}_n\}$ and search for the matrices **J** and **P** that minimize the functional

$$\min_{\substack{\mathbf{P}^T\mathbf{P} = \mathbf{I} \\ \mathbf{J} \in J}} f(\mathbf{P}, \mathbf{J}) = \min_{\substack{\mathbf{P}^T\mathbf{P} = \mathbf{I} \\ \mathbf{J} \in J}} \|\mathbf{U}\mathbf{P} - \mathbf{J}\|_F. \tag{22}$$

We follow the approach of Yu and Shi (2003) for the solution of this optimization problem, that iteratively computes this discretization using singular value decomposition and non-maximum supression. In what follows the clustering solutions obtained by employing k-means on the *p*-eigenvectors are denoted as pGrass-kmeans, and the ones obtained by the rotation of the normalized eigenvectors after solving (22) are denoted as pGrass-disc.

## 3.5 Multiway *p*-Grassmann clustering algorithm

---
**Algorithm 1** $p$-Grassmann spectral clustering
---
**Input:** adjacency matrix $\mathbf{W}$, number of clusters $k$, $\kappa$, $\theta$, final $p$ value $p_\omega$, normalized

**Output:** cluster indices $\mathbf{c}_{\text{best}}$, cut value $r_{\text{best}}$

  1  **function** pGrassmannClustering
  2      **if** normalized **then**
  3          Find $\mathbf{U}$: $\underset{\mathbf{U} \in \mathbb{R}^{(k,n)}}{\text{minimize}} \, F_2^{(n)}(\mathbf{U})$ using $\mathbf{W}$     ▷ See section 2.4
  4          Cut = NCut
  5      **else**
  6          Find $\mathbf{U}$: $\underset{\mathbf{U} \in \mathbb{R}^{(k,n)}}{\text{minimize}} \, F_2(\mathbf{U})$ using $\mathbf{W}$
  7          Cut = RCut
  8      **end if**
  9      $\mathbf{c}_{\text{best}} = \mathbf{c} = \text{discretize}(\mathbf{U})$     ▷ Obtain discrete solution; see section 3.4
10      $r_{\text{best}} = r_{\text{new}} = r_{\text{old}} = \text{Cut}(\mathbf{c})$     ▷ Initialize the cut value acc. (3) or acc. (4)
11      $p = 2$     ▷ Initialize the value of $p$.
12      **while** $p \geq p_w$ && $r_{\text{new}} \leq 1.05 \cdot r_{\text{old}}$ **do**
13          Reduce $p$     ▷ Pseudocontinuous reduction acc. (21)
14          **if** normalized **then**
15              Find $\mathbf{U}$: $\underset{\mathbf{U} \in \mathcal{G}r(k,n)}{\text{minimize}} \, F_p^{(n)}(\mathbf{U})$ using $\mathbf{W}$     ▷ See section 3.2.
16          **else**
17              Find $\mathbf{U}$: $\underset{\mathbf{U} \in \mathcal{G}r(k,n)}{\text{minimize}} \, F_p(\mathbf{U})$ using $\mathbf{W}$
18          **end if**
19          $\mathbf{c} = \text{discretize}(\mathbf{U})$     ▷ Obtain discrete solution; see section 3.4
20          $r_{\text{old}} = r_{\text{new}}$
21          $r_{\text{new}} = \text{Cut}(\mathbf{c})$     ▷ Update the cut value acc. (3) or acc. (4)
22          **if** $r_{\text{new}} < r_{\text{best}}$ **then**
23              $r_{\text{best}} = r_{\text{new}}$
24              $\mathbf{c}_{\text{best}} = \mathbf{c}$
25          **end if**
26      **end while**
27      **return** $\mathbf{c}_{\text{best}}$, $r_{\text{best}}$     ▷ optimal solution
28  **end function**
---

A general summary of the algorithmic scheme employed for the unnormalized (RCut based) or normalized (NCut based) multiway $p$-Grassmann spectral clustering is offered in Algorithm 1. The inputs of the algorithm are the adjacency matrix **W** of the graph in question, the number of the desired clusters $k$, the parameters $\kappa$, $\theta$ from (21), the final value of $p$ denoted as $p_w$, and whether the clustering will be based on the unnormalized (RCut) or the normalized (NCut) objective. As output we obtain the indices of the vertices forming the clusters and the discrete cut value obtained from the clustering. In steps 2–8 we solve the optimization problem for $p = 2$ and obtain $k$-eigenvectors stored in matrix **U**. Their discretization, through the k-means algorithm or the solution of (22) is performed in step 9, and the cut value is initialized accordingly in step 10. The main loop of the algorithm in steps 12–25 terminates if the value of $p$, which is initialized in 11 and reduced in a pseudocontinuous fashion in 13, reaches the final value $p_w$ or the cut value stops decreasing monotonically, with a tolerance of 5% on this monotonic reduction. The multiple unormalized or normalized $p$-eigenvectors are estimated on the manifold $\mathcal{G}r(k, n)$ in steps 14–18, and the discrete solution is obtained in 19. Then the cut values are updated in steps 20–24 if they are smaller than their value in the previous iteration.

## 4 Numerical results

We demonstrate in what follows the effectiveness of the $p$-Grassmann spectral clustering method, summarized in Algorithm 1. In Sect. 4.1 we report the setup of our numerical experiments and in Sect. 4.2 we outline the external methods considered in our comparisons, and discuss on the key differences between our approach and the most closely related method considered. Our results on synthetic graphs are presented in Sect. 4.3 and on graphs emerging from facial image and handwritten characters classification problems in Sect. 4.4.

### 4.1 Experimental setup

For all test cases we report results concerning the quality of the cut in terms of RCut (3) and NCut (4), unless specified otherwise. The corresponding accuracy of the labelling assignment is measured in terms of the unsupervised clustering accuracy (ACC $\in [0, 1]$) and the normalized mutual information (NMI $\in [0, 1]$) (Dalianis, 2018). For both metrics a value of 1 suggests a perfect grouping of the nodes according to the true labels. To this end, we work strictly with graphs that have ground-truth labels, and set the number of clusters $k$ equal to the total number of labelled classes. However, our approach is directly applicable to graphs with no ground-truth information for unsupervised community detection. We use MATLAB R2020a for our implementation, and run experiments on a total of 80 graphs, organized in 2 sets. The first one comprises 27 artificial test cases, with the purpose of demonstrating the impact of different optimization aspects of $p$-Grassmann spectral clustering. The second one includes 53 graphs originating from image classification and text recognition applications. For all methods under consideration we report the mean results after 10 runs. In all numerical experiments the connectivity matrix $\mathbf{G} \in \mathbb{R}^{n \times n}$ is created from a $k$-nearest neighbors routine, with the number of nearest neighbors (NN) set such that the resulting graph is connected. The similarity matrix $\mathbf{S} \in \mathbb{R}^{n \times n}$ between the data points is defined similarly to Zelnik-Manor and Perona (2005) as

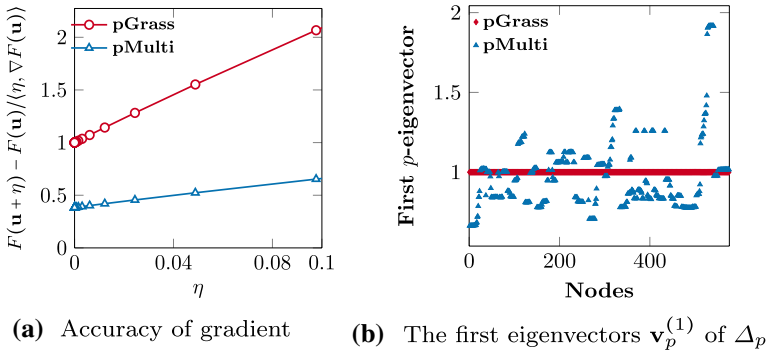

**(a)** Accuracy of gradient  **(b)** The first eigenvectors $\mathbf{v}_p^{(1)}$ of $\Delta_p$

**Fig. 4** Analysis of key differences between (Luo et al., 2010) and our work. **a** The accuracy of the approximated gradient used in Luo et al. (2010) compared against its numerical approximation using first order Taylor approximation. The x-axis denotes the different step size ($\eta$) used in the Taylor expansion (see text for details). The experiment is conducted using the UMIST dataset with a $p$ value of 1.8 and $k = 20$ number of clusters. **b** The values of the first eigenvector $\mathbf{v}_p^{(1)}$ of the graph $p$-Laplacian $\Delta_p$ for the UMIST dataset, estimated by the method in Luo et al. (2010) (pMulti) and by our approach (pGrass)

$$s_{ij} = \max\{s_i(j), s_j(i)\} \text{ with } s_i(j) = \exp\left(-4\frac{\|x_i - x_j\|^2}{\sigma_i^2}\right)$$

with $\sigma_i$ standing for the Euclidean distance between the $i$th data point and its $k$th nearest neighbor. The adjacency matrix $\mathbf{W}$ is then created as

$$\mathbf{W} = \mathbf{G} \odot \mathbf{S}. \tag{23}$$

The maximum number of Newton iterations for our method is set to 20 for every $p$-level, and the final $p$-level is set to $p_w = 1.1$. We fix the parameters of (21) at $\kappa = 0.9, \theta = 1.25$. This selection results in the following 8 total $p$-levels, i.e. $p = \{2, 1.9, 1.71, 1.539, 1.3851, 1.2466, 1.171, 1.1\}$. When using k-means for the discretization of the $p$-eigenvectors (pGrass-kmeans) we run k-means with 10 orthogonal and 20 random initial guesses. In order to select the best result out of the different k-means runs we use our objective, i.e., RCut or NCut (lower the better) as the primary ranking metric. Then the labelling accuracy metrics (ACC, NMI) are calculated based on the clusters obtained from the minimization of the cut values. When solving (22) for the clustering of the $p$-eigenvectors (pGrass-disc) the solution is unique.

### 4.2 Methods under consideration

We compare our method against a diverse selection of state-of-the-art clustering algorithms:

1. Spec (Luxburg, 2007): Traditional direct multiway spectral clustering. We consider the eigenvectors of the combinatorial graph Laplacian $\Delta_2$ for unnormalized clustering, and follow the approach of Yu and Shi (2003) with $\Delta_2^{(n)}$ for the normalized case.
2. pSpec (Bühler & Hein, 2009): Recursive bi-partitioning with the unnormalized and normalized graph $p$-Laplacian, using a hybrid Newton-gradient descent scheme for the minimization of the nonlinear objective.

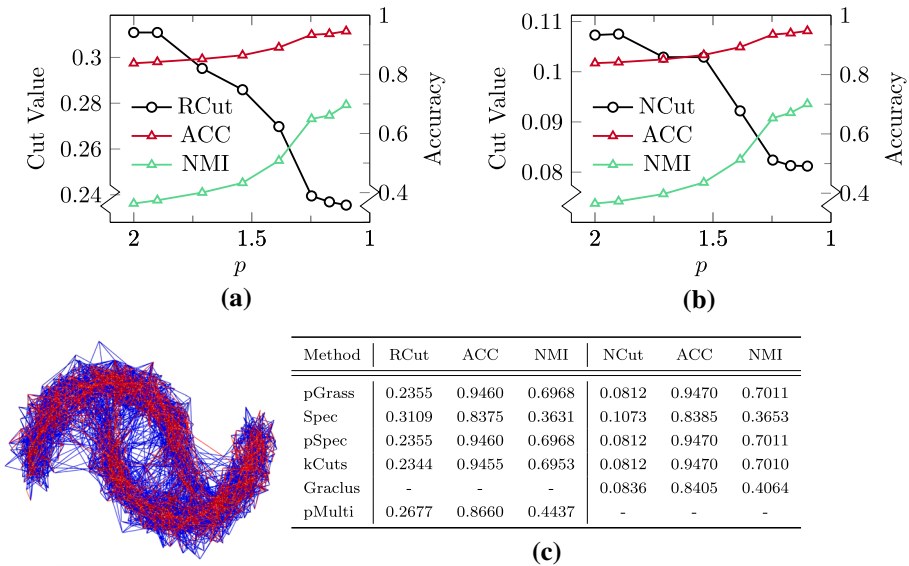

**Fig. 5** Clustering the two-moons dataset (illustrated). **a** The estimation of RCut and of the associated ACC and NMI with the pGrass algorithm for $p \in [1.1, 2]$. **b** The estimation of NCut and of the associated ACC and NMI with the pGrass algorithm for $p \in [1.1, 2]$. **c** Comparative results for the clustering methods under consideration

3. kCuts (Rangapuram et al., 2014): A tight continuous relaxation for the balanced direct $k$-cut problem, using a monotonically descending algorithm for the minimization of the resulting sum of Rayleigh quotients. The method is applicable for the minimization of a variety of discrete graph cut metrics, including RCut and NCut. We use 12 starting initializations for the routine, as suggested by the authors.
4. Graclus (Dhillon et al., 2007): A multilevel algorithm that optimizes for various weighted graph clustering objectives using a weighted kernel k-means objective, thus eliminating the need for eigenvector computations. We use the Kerhighan–Lin (Kernighan & Lin, 1970) algorithm at the coarsest level clustering and 10 local searches at each level for increased accuracy. Graclus minimizes directly only the NCut, thus it is omitted from any comparisons in the computation of the RCut and the associated accuracy metrics.
5. pMulti (Luo et al., 2010): The first full eigenvector analysis of $p$-Laplacian leading to direct multiway clustering, and the most directly related method to our $p$-Grassmann approach. The discrete minimization objective for this approach is the RCut, thus we omit it from any NCut based comparisons.

The code for the methods outlined in 1–4 is available online.[3] We implement method 5, as described in Luo et al. (2010), and briefly outline here the key differences from our approach. The minimization of the constrained multiway $p$-spectral problem (14) is achieved through an approximated gradient descent scheme which suffers from inaccuracies. This is illustrated in Fig. 4a where the ratio of directional derivative $F'$ obtained using a first order Taylor expansion[4] is compared to that of the computed gradient from Luo et al. (2010) and ours (19), for the UMIST dataset (Graham & Allinson, 1998) at $p = 1.8$. The ratio of $(F(\mathbf{u} + \boldsymbol{\eta}) - F(\mathbf{u}))/\langle \boldsymbol{\eta}, \nabla F(\mathbf{u}) \rangle$ should ideally approach one as the step size $\eta$ in the Taylor expansion decreases. However, with the approximated gradient defined in Luo et al. (2010) this is not the case (see Fig. 4a). Due to this gradient inaccuracy, fundamental properties of the spectrum of $\boldsymbol{\Delta}_p$ are no longer valid. For example, the degeneracy of the eigenvalues, corresponding to the constant eigenvectors $\mathbf{v} = c \cdot \mathbf{e}$, no longer indicates the number of connected components in the graph. In contrast, our $p$-Grassmann approach, referred to as pGrass, preserves this fundamental property of $\boldsymbol{\Delta}_p$, as illustrated in Fig. 4b. Furthermore, since the functional $F$ is nonconvex the modified gradient descent approach used in their work has a suboptimal convergence rate, as opposed to the properties of our method. Finally, the linear reduction rate of $p$ in Luo et al. (2010) results in fewer evaluations taking place close to $p \approx 1$, and their method consider only the minimization of the unnormalized $p$-spectral objective, associated with the RCut metric.

---

[3] The Spec code is available at: https://github.com/panji530/Ncut9. The pSpec code is available at: https://www.ml.uni-saarland.de/code/pSpectralClustering. The kCuts code is available at: https://www.ml.uni-saarland.de/code/balancedKCuts. The Graclus code is available at: https://www.cs.utexas.edu/users/dml/Software/graclus.html.

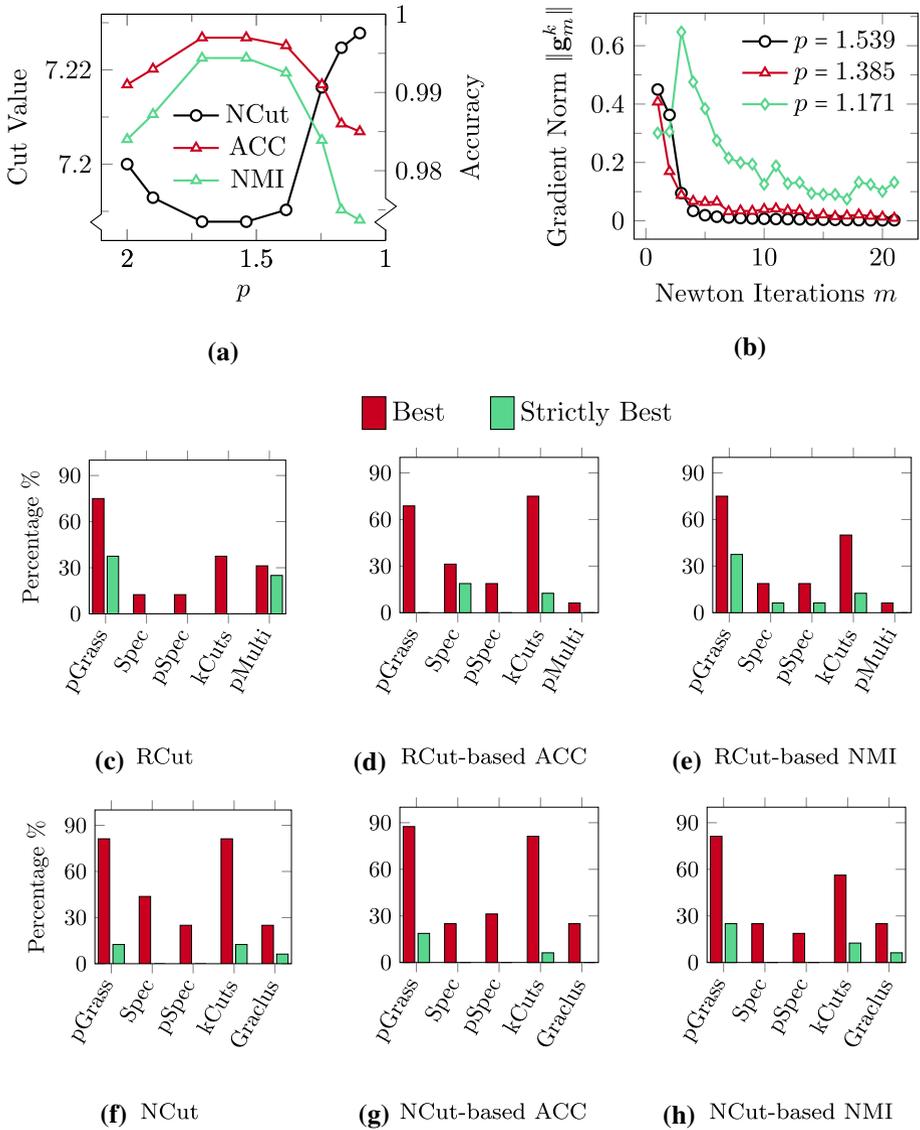

**Fig. 6** Clustering the LFR benchmark datasets with a noise component $\mu$. **a** NCut values and the associated ACC and NMI for $\mu = 0.38$ for a decreasing value of $p$. **b** Norm of the gradient $\|\mathbf{g}_m^k\|$ over Newton iterations $m$ for $\mu = 0.38$ for three different $p$-levels. **c–h** Collective results of the fraction of times a method achieves the best and the strictly best metrics for the entire benchmark with $\mu \in [0.1, 0.4]$

### 4.3 Artificial datasets experiments

In this subsection we focus on artificial datasets widely used as test cases for clustering algorithms, in order to display the behavior of our pGrassmann clustering algorithm in challenging scenarios. In particular, in Sect. 4.3.1 we are interested in studying the effect that the reduction of the value of $p$ has on the clustering result for a graph corrupted by high-dimensional noise and for a set of 16 stochastic block model graphs. In Sect. 4.3.2, we shift our attention to Gaussian datasets, and study the impact of a large number of ground-truth classes on the accuracy of our method. Last, in Sect. 4.3.3 we take a closer look at the eigenvectors of the graph $p$-Laplacian and the differences between standard spectral and $p$-spectral embedding on a synthetic dataset with three ground-truth classes. The results obtained by discretizing the $p$-eigenvectors with the k-means algorithm and by solving (22) are almost identical for these artificial datasets, therefore in what follows in this subsection only the results of pGrass-kmeans are presented, and are referred to as pGrass.

### 4.3.1 Reducing the value of *p*

We initially study the impact of the reduction of the value of $p \in (1, 2]$ in (18) on the high-dimensional two-moons dataset, which is commonly used in evaluating graph clustering algorithms. It consists of two half-circles in $\mathbb{R}^2$ embedded into a 100-dimensional space with Gaussian noise $\mathcal{N}(0, \sigma^2 \mathbb{I}_{100})$. This high-dimensional noise results in a complex edge formation, as illustrated in Fig. 5 for $n = 2000$ points and a variance of $\sigma^2 = 0.02$. In Fig. 5a we show the effect of reducing $p$ from 2 towards 1 on the resulting RCut and on the associated labelling accuracy metrics (ACC, NMI), and in Fig. 5b we show the accuracy results of the normalized $p$-Grassmann clustering variant with NCut as its objective. In both cases the monotonic descent of the graph cut metrics leads to nearly perfect accuracies at $p = 1.1$. In Fig. 5c we present the results obtained by all the methods considered. Our algorithm performs significantly better than Spec, pMulti and Graclus, while it achieves almost identical cut and accuracy values to the pSpec and kCuts methods. The identical results with pSpec are expected in this case, as for a number of clusters $k = 2$ the minimization objective of Bühler and Hein (2009) is equivalent with ours (18).

---

[4] The first order Taylor expansion reads $F(\mathbf{u} + \boldsymbol{\eta}) = F(\mathbf{u}) + \langle \boldsymbol{\eta}, \nabla F(\mathbf{u}) \rangle$, where $\eta$ is the step size.

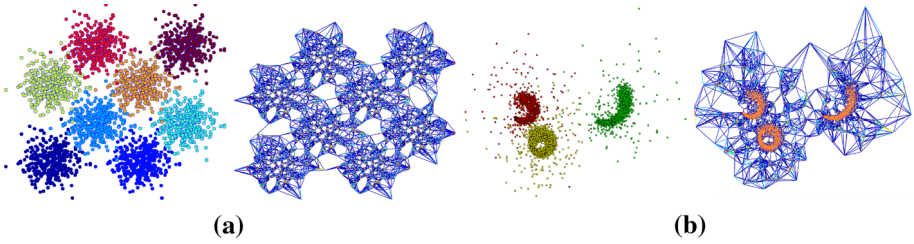

**Fig. 7** A subset of the synthetic datasets used. **a** One of the Gaussian datasets considered in Sect. 4.3.2 with $k = 8$ gound-truth clusters, $n = 3200$ nodes and $m = 19319$ edges. **b** The worms dataset, considered in Sect. 4.3.3, consists of $n = 5967$ points with three ground-truth communities. The resulting graph has $m = 36031$ edges. (Best viewed in color.)

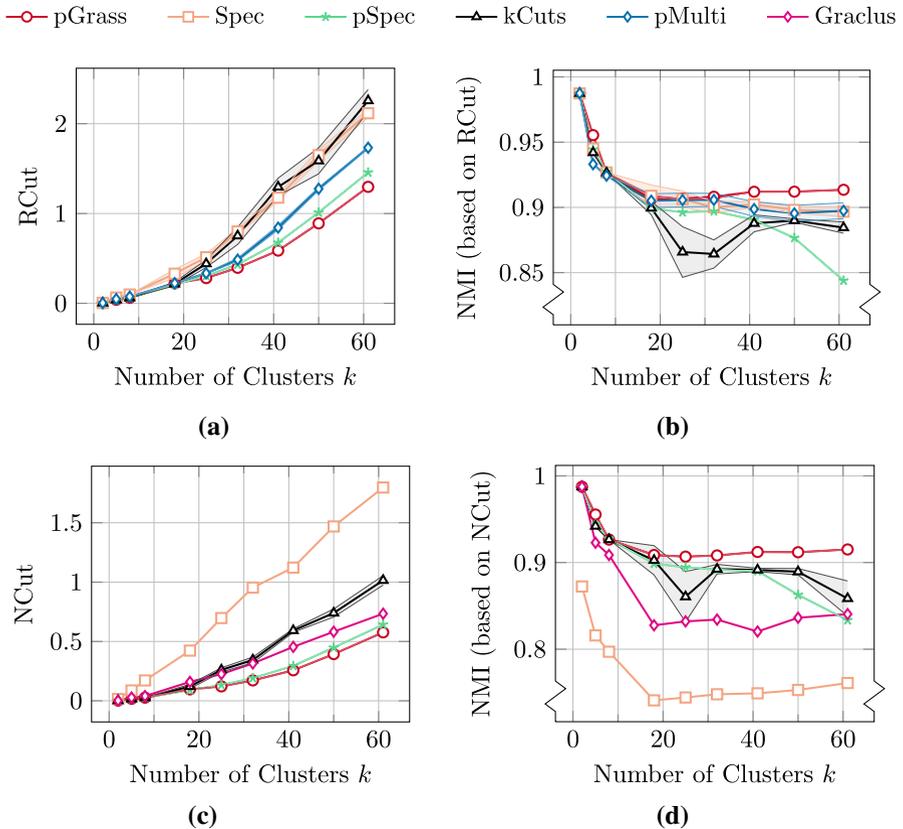

**Fig. 8** Clustering the Gaussian datasets with an increasing number of clusters $k$. **a** RCut values for all the methods under consideration. **b** NMI values for all the methods under consideration based on the RCut. **c** NCut values for all the methods under consideration. **d** NMI values for all the methods under consideration based on the NCut

We further demonstrate in Fig. 6 the effectiveness of the introduced Algorithm 1 in finding the best available clusters even in scenarios where the discrete graph cut metrics are not monotonically descending for a decreasing value of $p$. To this end, we consider the LFR model (Lancichinetti et al., 2008), which is a stochastic block model whose nodes' degrees follow the power law distribution with a parameter $\mu$ controlling what fraction of a node's neighbours is outside the node's block. We follow the approach of Fountoulakis et al. (2020) and pick $\mu \in [0.1, 0.4]$, with this range giving rise to graphs that contain increasingly noisy clusters for an increasing value of $\mu$. The number of clusters in this benchmark ranges from $k = 17$ to $k = 20$. In Fig. 6a we show the value of NCut and of the associated accuracy metrics ACC and NMI for the case with $\mu = 0.38$. The monotonic minimization of NCut, with a tolerance of 5% as specified in Algorithm 1, is interrupted at $p = 1.539$. At this $p$-level the graph cut reaches a minimum value of NCut = 7.187, with the corresponding accuracy metrics being at their maximum values ACC = 0.9970, NMI = 0.9944. Our algorithm stops the reduction of the value of $p$ at this level, however we report the results of the optimization procedure up to the final level of $p = 1.1$ in order to demonstrate that the increasing nonlinearity close to $p \approx 1$ may lead to unfavorable results. At the final $p$-level the value of the graph cut has ascended to NCut = 7.228, with the values of the accuracy metrics being decreased at ACC = 0.9850 and NMI = 0.9737. In Fig. 6b we plot the norm of the gradient $\|\mathbf{g}_m^k\|$ over the Newton iterations $m$ for the levels $p = 1.539$ (best solution) and two subsequent levels closer to $p \approx 1$ ($p = 1.385, p = 1.171$). The monotonic minimization of $\|\mathbf{g}_m^k\|$ at the best $p$-level is followed by an increasingly oscillating behavior as $p \to 1$. This showcases that the monotonic minimization of our discrete graph cut metric NCut is directly associated with the monotonic decrease of the gradient norm of our continuous objective (19). The proposed Algorithm 1 is guaranteed to find the best available solution from all $p$-levels under consideration. This is highlighted in Fig. 6c–h, where the results for all the LFR benchmark datasets (in total 16 cases) for all the methods under consideration are collected. We present the percentage of times a method found the best and the strictly best solution in terms of graph cut metrics (RCut, NCut), and the associated labelling accuracy values in ACC and NMI. Our $p$-Grassmann clustering routine outperforms the external methods Spec, pSpec, pMulti and Graclus in all the metrics under question, and achieves comparable scores with the kCuts algorithm. In particular, the unnormalized pGrass algorithm achieves the best—strictly best solutions in 75–37.5% of the cases when minimizing the RCut, in 68.75–0% when finding the associated ACC and in 75–37.5% when finding the associated NMI. The corresponding percentages for the kCuts algorithm are 37.5–0% for RCut, 75–12.5% for ACC, and 50–12.5% for NMI. The normalized pGrass algorithm achieves the best—strictly best solutions in 81.25–12.50% of the cases when minimizing the NCut, in 87.50–18.75% when finding the associated ACC, and in 81.25–25% when finding the associated NMI. The corresponding percentages for the kCuts algorithm are 81.25–12.5% for NCut, 81.25–6.25% for ACC, and 56.25–12.5% for NMI. The numeric values of the results for the LFR benchmark datasets are summarized in Table 2 in "Appendix B".

### 4.3.2 Increasing the number of clusters (*k*)

In order to study the clustering quality of our algorithm as the number of clusters ($k$) increases we utilize a set of synthetic Gaussian datasets with an increasing number of ground-truth clusters. Each dataset consists of $k$ clusters containing 400 points each. The clusters are generated using a Gaussian distribution with a variance of $\sigma^2 = 0.055$, with the mean of each cluster then placed equidistantly on a 2D square grid (see Fig. 7a). For the experiment we generated datasets with varying $k = \{2, 5, 8, 18, 25, 32, 41, 50, 61\}$, resulting in 9 graphs with an increasing number of nodes, edges and clusters.

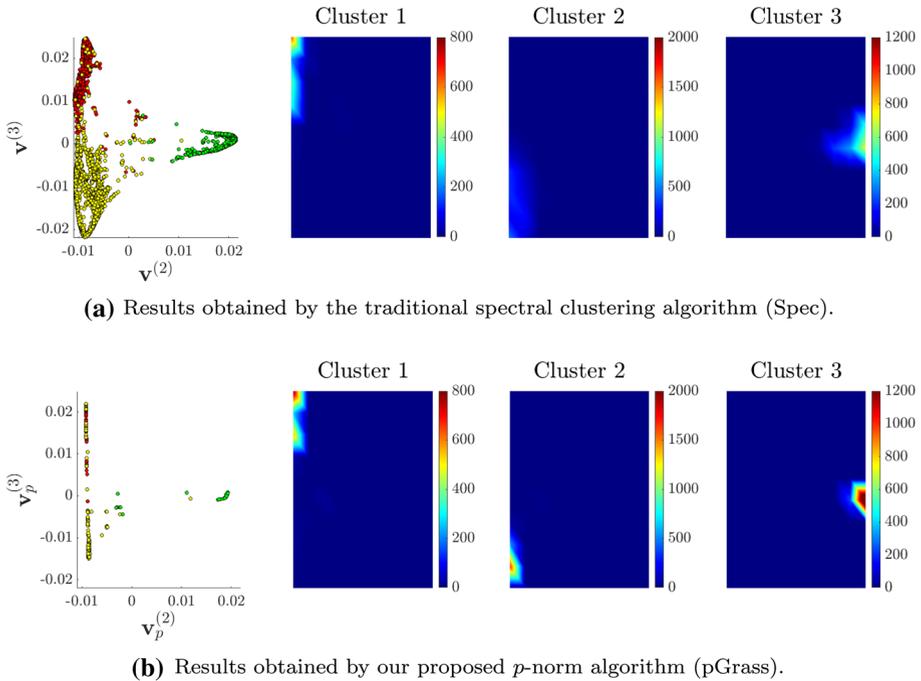

**Fig. 9** Embedding results for the worms dataset. Starting from the left, the points of the dataset are illustrated using the entries of the second and third eigenvectors of the $\Delta_2$ in (**a**), and of $\Delta_p$ for $p = 1.1$ in (**b**), as $x$ and $y$ coordinates. The heat maps that follow depict the density of the points from each of the three clusters. (Best viewed in color.)

In Fig. 8 we present the mean values and the standard deviation of the cut metrics and the associated accuracy metric NMI for the Gaussian datasets. In Fig. 8a we show the results obtained when minimizing RCut and in Fig. 8b the corresponding NMI. Our pGrass clustering routine finds the minimum RCut in 7/9 cases and the strictly minimum in 5/9 cases. In terms of NMI, pGrass attains the maximum in 8/9 cases and the strictly maximum in 7/9 cases. The results obtained when attempting to minimize the NCut are shown in Fig. 8c, with the corresponding NMI values shown in Fig. 8d. Our algorithm finds the best NCut in 7/9 cases and the strictly best in 5/9 cases. In terms of NMI our algorithm fares the best in 7/9 cases and the strictly best in 6/9 cases. No significant deviations from the mean values are reported for pGrass.

We note that in both the normalized and unnormalized experiments the benefits of our method are becoming more evident as the number of clusters $k$ increases. In particular, for $k \geq 25$ pGrass attains the strictly best results in terms of both graph cut and labelling accuracy. This behavior demonstrates that the $p$-Grassmann algorithm is favorable for clustering datasets with a large number of clusters not only from recursive approaches

Table 1 Clustering results for the facial image datasets of Sect. 4.4.1

| Method | Olivetti | | | Faces95 | | | FACES | | |
|---|---|---|---|---|---|---|---|---|---|
| | NCut | ACC | NMI | NCut | ACC | NMI | NCut | ACC | NMI |
| pGrass—kmeans | **3.984** | −4.15% | −2.28% | **2.658** | −5.77% | −4.24% | **29.42** | −3.58% | −2.41% |
| pGrass—disc | −4.50% | **0.716** | **0.831** | −4.50% | **0.609** | **0.758** | −6.08% | **0.802** | **0.91** |
| Spec (%) | −24.84 | −9.19 | −5.27 | −24.84 | −4.23 | −0.90 | −15.05 | −2.50 | −1.23 |
| pSpec (%) | −8.04 | −7.41 | −3.06 | −8.04 | −6.86 | −6.02 | −4.34 | −6.73 | −2.71 |
| kCuts (%) | −1.41 | −6.78 | −3.20 | −1.41 | −10.37 | −7.70 | −7.67 | −13.0 | −6.99 |
| Graclus (%) | −23.10 | −6.36 | −2.25 | −23.11 | −9.25 | −2.38 | −9.98 | −3.70 | −2.56 |

Both variants of our algorithm, pGrass-kmeans and pGrass-disc are considered. For each dataset we report the mean value (in bold) of NCut, ACC and NMI achieved by the best method, and the percentage the remaining methods are inferior to that value

(pSpec, Graclus), which is expected due to the recursive methods' instabilities (Simon & Teng, 1997), but also from the direct multiway methods under consideration (Spec, kCuts, pMulti).

### 4.3.3 *p*-spectral embedding.

In order to highlight the differences between the embeddings achieved using the eigenvectors of the combinatorial graph Laplacian $\Delta_2$ and those of the graph *p*-Laplacian $\Delta_p$, we utilize the Worms2 dataset (Sieranoja & Fränti, 2019). This dataset is composed of three individual worm-like shapes that start from a random position and move towards a random direction. Points are drawn according to a Gaussian distribution with both low and high variance components that are gradually increasing as the points populate the 2D space. The direction of the generation of each worm-like shape is orthogonal to the previous one. The dataset consists of $n = 5967$ points with three ground-truth communities and the resulting graph has $m = 36{,}031$ edges (see Fig. 7b).

We visualize the embedding results obtained by standard spectral clustering (Spec) and our method (pGrass) in Fig. 9. There are three distinct clusters in the dataset. We utilize the second and third eigenvectors as the *x*- and *y*-axis, respectively. The *p*-spectral embedding (Fig. 9b) organizes the nodes of the dataset in clearly distinguishable groups, as opposed to the spectral embedding (Fig. 9a). The heat maps illustrate the density of points from each cluster in the two different embeddings ($p = 2, p = 1.2$). We consider ten bins for each direction in order to measure the density for each cluster. The limits of the colorbar are set in both cases to the maximum density values obtained by our method, for a clear comparison.

Upon visual inspection, the pGrass algorithm performs superior to its the traditional Spec routine in the task of creating sharp cuts of the data. Since the last stage of both algorithms is to cluster these points according to their relative distances (see Sect. 3.4 on discretization), the *p*-spectral coordinates of Fig. 9b are expected to lead to clusters of higher quality. This hypothesis is supported by our numerical results. The quality of the cut achieved by our algorithm (RCut = 0.0062) is 49.6% better compared to the one obtained by spectral clustering (RCut = 0.0123). This improvement in terms of graph cut criteria

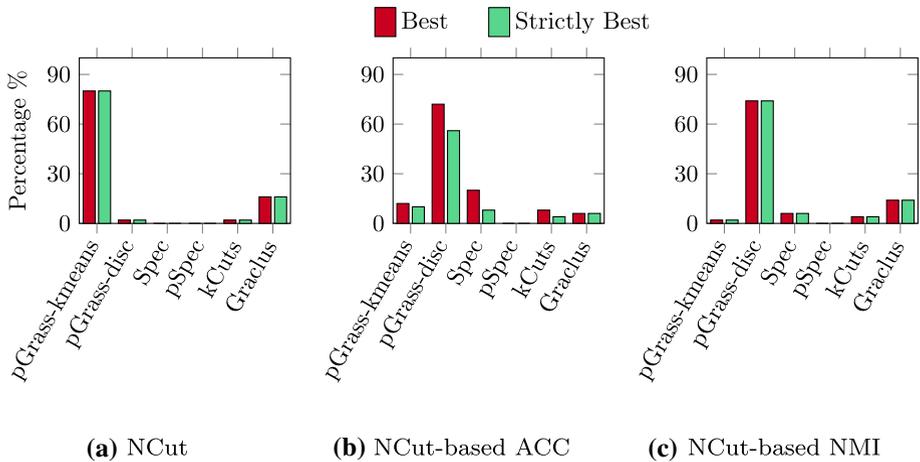

**Fig. 10** Clustering the Omniglot database of handwritten digits. The red bar indicated the percentage of times that a method achieved the best solution and the green bar the percentage of times it achieved the strictly best solution. **a** NCut values, **b** ACC values based on NCut, **c** NMI values based on NCut

also leads to a slightly better labelling accuracy for this dataset, with our method achieving scores of $ACC = 0.985, NMI = 0.93$ as opposed to the traditional spectral routine which achieves scores of $ACC = 0.981, NMI = 0.92$.

### 4.4 Real-world experiments

We now proceed with the application of our $p$-Grassmann spectral clustering, as outlined in Algorithm 1, in graphs emerging from real-world problems. In Sect. 4.4.1 we consider the problem of classifying facial images according to their labels and in Sect. 4.4.2 the problem of distinguishing handwritten characters. In both cases graphs are created according to the procedure outlined in Sect. 4.1. We report the mean results, obtained after 10 runs for each method under consideration, in minimizing the NCut and the corresponding labelling accuracies. We present here the both the results obtained after applying k-means for the clustering of the $p$-eigenvectors (pGrass-kmeans), and the results after solving (22) (pGrass-disc).

#### 4.4.1 Classification of facial images

We consider the following publicly available[5] datasets depicting facial expressions

- Olivetti (Samaria & Harter, 1994): A set of 10 different facial images of 40 distinct subjects at resolution $64 \times 64$ pixels, taken at different times, varying lighting, facial expressions and facial details.

---

[5] The Olivetti dataset is available at https://cam-orl.co.uk/facedatabase.html. The faces95 is available at https://cmp.felk.cvut.cz/spacelib/faces/. The FACES dataset is available after registration at https://faces.mpdl.mpg.de/imeji/.

- Faces95 (Hond & Spacek, 1997): A collection of 1440 pictures with resolution $180 \times 200$ pixels from 72 individuals that were asked to move while a sequence of 20 images was taken.
- FACES (Ebner et al., 2010): A set of images with resolution $2835 \times 3543$ pixels of naturalistic faces of 171 individuals displaying 6 facial expressions. The database consists of 2 sets of pictures per person and per facial expression, resulting in a total of 2052 images. We downsample the images at 20% of their initial resolution to decrease the problem size when creating the adjacency matrix **W**.

For these datasets the number of nearest neighbors needed for a connected graph is NN = 6 for Olivetti faces and NN = 10 for both Faces95 and FACES. We summarize our results in Table 1. For each dataset we report the mean value (in bold) of NCut, ACC and NMI achieved by the best method, and the percentage the remaining methods are inferior to that value. Inferiority in percentage values is defined as $I = 100 \cdot \gamma \cdot (e_{\text{ref}} - e_{\text{best}})/e_{\text{best}}$, where $e_{\text{best}}$ is the best value, $e_{\text{ref}}$ the value it is compared against, and $\gamma = -1$ for minimization scenarios (NCut) and $\gamma = 1$ for maximization ones (ACC, NMI). Our algorithmic variant pGrass-kmeans finds the best NCut result in all three datasets. However, these minimum cut values do not correspond to a maximization of the labelling accuracy metrics. Instead, our algorithmic variant pGrass-disc, which discretizes the eigenvectors of the normalized graph $p$-Laplacian $\Delta_p^{(n)}$ with the orthonormal transformation described in Yu and Shi (2003), achieves the highest ACC and NMI values in all cases. Similarly to the numerical experiments on artificial datasets of Sect. 4.3.2, no significant deviations ($< 1\%$) from the mean reported values are observed for pGrass.

### 4.4.2 Classification of handwritten characters

For the problem of classifying handwritten characters we consider the Omniglot database[6] (Lake et al., 2015). It consists of 1623 different handwritten characters from 50 alphabets. Each of the 1623 characters was drawn online via Amazon's Mechanical Turk by 20 different people, with each drawing having dimensions of $105 \times 105$ pixels. For each of these 50 alphabets we consider the problem of grouping the symbols in their respective classes. The number of nearest neighbors is set to NN = 10 for all cases.

We present the percentage of times a method achieved the best and the strictly best solution in Fig. 10. In Fig. 10a we see that our variant pGrass-kmeans obtains the best NCut values in 80% of the cases, with the remaining methods pGrass-disc, Spec, pSpec, kCuts and Graclus obtaining 2%, 0%, 0%, 2%, and 16% respectively. There are no ties in the NCut results, thus best and strictly best percentages are identical. In terms of ACC, illustrated in Fig. 10b, our variant pGrass-disc find the best solution in 72% of the cases, and the strictly best in 56%. The remaining methods pGrass-kmeans, Spec, pSpec, kCuts, Graclus find the best-strictly best solution in 12–10%, 20–8%, 0–0%, 8.0–4.0% and 6.0–6.0% of the cases respectively. Finally, the NMI results of Fig. 10c indicate that the pGrass-disc variant finds the best solution in 74% of the cases, with the remaining methods pGrass-kmeans, Spec, pSpec, kCuts, Graclus achieving a score of 2%, 6%, 0%, 4%, and 14% respectively. Similarly to the NCut results all NMI solutions are unique, thus best and strictly best percentages are identical. The numeric values of the results for the Omniglot database are summarized in Table 3 in "Appendix B".

---

[6] The Omniglot database is available at https://github.com/brendenlake/omniglot.

### 4.5 Discussion of real-world results

All real-world numerical experiments presented above demonstrate that clustering with the pGrass algorithm leads to either obtaining the best (minimum) graph cut values, or the best (maximum) labelling accuracy metrics. In particular, the pGrass-kmeans variant, that discretizes the resulting $p$-eigenvectors from the Grassmannian optimization problem (18) with the k-means algorithm, showcases superior results in terms of the balanced graph cut metric NCut (4). This variant (pGrass-kmeans) attains the minimum NCut value from all the methods under consideration in all three facial expression datasets under question in Sect. 4.4.1, and the minimum NCut value in 80% of the total 50 handwritten datasets of Sect. 4.4.2. All these solutions were unique, i.e., none of the external graph clustering methods under consideration reported the same cut. However, as reported in multiple related works (Bühler & Hein, 2009; Rangapuram et al., 2014; Fountoulakis et al., 2020) the minimization of balanced graph cut metrics does not necessarily lead to a increase in the accuracy of the labelling assignment for real-world data. The creation of the adjacency matrix plays a vital role in this discrepancy, and is an active field of research (Egilmez et al., 2017; Kumar et al., 2020; Slawski and Hein 2015). We demonstrate that for widely used adjacency matrices (23), employing a different technique for the discretization of the $p$-eigenvectors (pGrass-disc) leads to favorable labelling accuracy assignments, even if the graph cut values are inferior than the ones by pGrass-kmeans. Rotating the eigenvectors in order to obtain discrete partitions has been reported to be particularly successful in maximizing the clustering accuracy metrics of labelled image data (Zelnik-Manor & Perona, 2005; Zhu et al., 2020), and our numerical experiments further support this observation. The algorithmic variant pGrass-disc minimizes (22) as suggested in Yu and Shi (2003), and results in the maximum ACC and NMI for all three facial expressions datasets. In the classification of handwritten digits this variant finds the best ACC in 72%, the strictly best ACC in 56%, and the best NMI in 74% of the cases, with the NMI solutions being unique. This showcases that the $p$-spectral embeddings found by (18) can lead to the minimization of the balanced graph cut metric (pGrass-kmeans), which is the primary objective of graph partitioning applications, or to the maximization of the labelling assignment accuracy (pGrass-disc), which is the goal in classification problems.

## 5 Conclusions

In this work, we developed a new method for multiway $p$-spectral clustering that leverages recent advancements in Riemannian optimization. This was achieved by reformulating the problem of obtaining multiple eigenvectors of the graph $p$-Laplacian as an unconstrained minimization problem on a Grassmann manifold. Our method reduces the value of $p$ in a pseudo-continuous manner, and ensures that the best result with respect to balanced graph cut metrics is retrieved from the various $p$-levels. Our method is compared against various state-of-the-art methods in a series of artificial and real-world tests, showcasing that the introduced algorithm offers significant gains in the majority of the cases in terms of the balanced graph cut metrics and in terms of the accuracy of the labelling assignment. For the high-dimensional two moons example we find that pGrass improves monototically the discrete graph cut metrics and the labelling accuracies as the value of $p$ tends to 1. Here pGrass performs equivalently to other $p$-spectral routines and outperforms traditional clustering methods. When clustering various graphs from the stochastic block model we observe that the monotonic minimization of the discrete metrics is not guaranteed close to $p \approx 1$, but pGrass selects the best solution from the available $p$-levels. Here our algorithm fares the best in the majority of the cases, and is comparable only with the external method kCuts. When testing Gaussian datasets, both the unnormalized and normalized pGrass clustering routines perform the best in all the metrics considered, with the benefits becoming more evident as the number of clusters increase. Furthermore, we saw how the spectral embeddings obtained by pGrass are more likely to lead to high quality clusterings than the embeddings achieved by traditional spectral methods in the 2-norm. Our results from the numerical experiments on real-world datasets highlight that the eigenvectors obtained by pGrass can lead to either superior graph cut values or labelling accuracy metrics, depending on the clustering method that transforms them into discrete partitions. The consistency of these results, from the artificial tests to the real-world cases, highlights the effectiveness of the introduced clustering algorithm and the broad applicability of the presented work.

## Appendix A

In this Appendix we show the derivation of the Euclidean gradient $g_m^k$ (introduced in (19)) and the approximate Hessian $h_{mn}^k$ (introduced in (20)) of the functional $F_p$.

The $m$-th entry of the Euclidean gradient ($\mathbf{g}^k$) of $F_p$ with respect to $u^k$ is

$$g_m^k = \frac{\partial F_p}{\partial u_m^k} = \frac{\partial}{\partial u_m^k} \sum_{i,j=1}^{n} \frac{w_{ij}\left|u_i^k - u_j^k\right|^p}{2\|\mathbf{u^k}\|_p^p} = \frac{\partial}{\partial u_m^k}\frac{A}{B},$$

$$g_m^k = \frac{\partial F_p}{\partial u_m^k} = \frac{\partial}{\partial u_m^k} \sum_{i,j=1}^{n} \frac{w_{ij}\left|u_i^k - u_j^k\right|^p}{2\|\mathbf{u^k}\|_p^p} \text{ where } A = \sum_{i,j=1}^{n} w_{ij}\left|u_i^k - u_j^k\right|^p \text{ and } B = 2\|\mathbf{u^k}\|_p^p.$$

$$g_m^k = \frac{1}{B}\frac{\partial A}{\partial u_m^k} - \frac{\partial B}{\partial u_m^k}\frac{A}{B^2} = \frac{1}{B}\left[\frac{\partial A}{\partial u_m^k} - \frac{\partial B}{\partial u_m^k}\frac{A}{B}\right], \text{ and applying the product rule} \quad (24)$$

$$\frac{\partial A}{\partial u_m^k} = \frac{\partial}{\partial u_m^k} \sum_{i,j=1}^{n} w_{ij}\left|u_i^k - u_j^k\right|^p = \sum_{i,j=1}^{n} w_{ij}p\left|u_i^k - u_j^k\right|^{p-1}\text{sign}(u_i^k - u_j^k)\frac{\partial}{\partial u_m^k}(u_i^k - u_j^k)$$

$$\frac{\partial A}{\partial u_m^k} = p\sum_{i,j=1}^{n} w_{ij}\phi_p(u_i^k - u_j^k)\frac{\partial}{\partial u_m^k}(u_i^k - u_j^k) \text{ , since } \phi_p(x) = |x|^{p-1}\text{ sign}(x), \text{ with} \quad (25)$$

$$\frac{\partial}{\partial u_m^k}(u_i^k - u_j^k) = \begin{cases} 1 & \text{if } i = m \text{ and } j \neq m \\ -1 & \text{if } j = m \text{ and } i \neq m \\ 0 & \text{else} \end{cases}. \quad (26)$$

Using (26) in (25) we get,

$$\frac{\partial A}{\partial u_m^k} = p \sum_{j=1}^n w_{mj}\phi_p(u_m^k - u_j^k) - p \sum_{i=1}^n w_{im}\phi_p(u_i^k - u_m^k)$$

$$\frac{\partial A}{\partial u_m^k} = p \sum_{j=1}^n \left[ w_{mj}\phi_p(u_m^k - u_j^k) - w_{jm}\phi_p(u_j^k - u_m^k) \right] \quad (27)$$

$$\frac{\partial A}{\partial u_m^k} = 2p \sum_{j=1}^n w_{mj}\phi_p(u_m^k - u_j^k), \text{ since } \phi_p(-x) = -\phi_p(x) \text{ and } w_{mj} = w_{jm}.$$

$$\frac{\partial B}{\partial u_m^k} = \frac{\partial}{\partial u_m^k} 2\|\mathbf{u^k}\|_p^p = \frac{\partial}{\partial u_m^k} 2\sum_{i=1}^n |u_i^k|^p = 2\sum_{i=1}^n p|u_i^k|^{p-1}\text{sign}(u_i^k)\frac{\partial u_i^k}{\partial u_m^k}$$

$$\frac{\partial B}{\partial u_m^k} = 2p\phi_p(u_m^k), \text{ since } \phi_p(x) = |x|^{p-1}\text{sign}(x) \text{ and } \frac{\partial u_i^k}{\partial u_m^k} = \left\{ \begin{array}{l} 1 \text{ if } i = m \\ 0 \text{ else} \end{array} \right\}. \quad (28)$$

Substituting (27) and (28) in (24) we get,

$$g_m^k = \frac{p}{\|\mathbf{u^k}\|_p^p}\left[ \sum_{j=1}^n w_{mj}\phi_p(u_m^k - u_j^k) - \phi_p(u_m^k) \sum_{i,j=1}^n \frac{w_{ij}\left|u_i^k - u_j^k\right|^p}{2\|\mathbf{u^k}\|_p^p} \right]. \quad (29)$$

The Euclidean Hessian of $F_p$ with respect to $u^k$ is the matrix $\mathbf{H^k}$. Its $m$-th row and $l$-th column entry is

$$h_{ml}^k = \frac{\partial g_m^k}{\partial u_l^k} = \frac{\partial}{\partial u_l^k} \frac{p}{\|\mathbf{u^k}\|_p^p}\left[ \sum_{j=1}^n w_{mj}\phi_p(u_m^k - u_j^k) - \phi_p(u_m^k) \sum_{i,j=1}^n \frac{w_{ij}\left|u_i^k - u_j^k\right|^p}{2\|\mathbf{u^k}\|_p^p} \right]. \quad (30)$$

The Hessian matrix as in (30) is not sparse and will cause storage problems. Therefore, we neglect the lower rank updates (refer to Bühler and Hein (2009) for details). The existing higher rank term can then be simplified, in the same way with the gradient derivation seen above, to arrive at the approximated Hessian:

$$\frac{\partial g_m^k}{\partial u_l^k} \approx \begin{cases} \frac{p(p-1)}{\|\mathbf{u^k}\|_p^p}\sum_{j=1}^n w_{mj}|u_m^k - u_j^k|^{p-2} & \text{if } m = l, \\ \frac{-p(p-1)}{\|\mathbf{u^k}\|_p^p} w_{ml}|u_m^k - u_l^k|^{p-2} & \text{otherwise.} \end{cases} \quad (31)$$

## Appendix B

We present in this Appendix the numerical results for the experiments of Sects. 4.3.1 and 4.4.2 on the LFR benchmark datasets and the Omniglot database of handwritten characters respectively. For the number of times each method under consideration found the best and the strictly best solutions we refer to Figs. 6 and 10 for the LFR datasets and the Omniglot cases respectively. The clustering methods under consideration are listed in Sect. 4.2 (see Tables 2, 3).

Table 2 Clustering results for the LFR benchmark datasets with an increasing noise component $\mu \in [0.1, 0.4]$

| Case | Measure | pGrass | Spec | pSpec | kCuts | Luo | Measure | pGrass | Spec | pSpec | kCuts | Graclus |
|---|---|---|---|---|---|---|---|---|---|---|---|---|
| $\mu = 0.10$ | RCut | **18.82** | **18.82** | − 0.1% | **18.82** | − 13.4% | NCut | **1.92** | **1.92** | − 0.75% | **1.92** | − 37.2% |
| | ACC | **1.0** | **1.0** | − 0.1% | **1.0** | − 10.6% | ACC | **1.0** | **1.0** | − 0.1% | **1.0** | − 5.3% |
| | NMI | **1.0** | **1.0** | − 0.2% | **1.0** | − 2.6% | NMI | **1.0** | **1.0** | − 0.16% | **1.0** | − 4.0% |
| $\mu = 0.12$ | RCut | **21.6** | **21.6** | **21.6** | **21.6** | − 17.89% | NCut | **2.18** | **2.18** | − 1.1% | **2.18** | − 19.3% |
| | ACC | **1.0** | **1.0** | **1.0** | **1.0** | − 6.3% | ACC | **1.0** | **1.0** | − 0.2% | **1.0** | − 2.7% |
| | NMI | **1.0** | **1.0** | **1.0** | **1.0** | − 2.4% | NMI | **1.0** | **1.0** | − 0.34% | **1.0** | − 2.0% |
| $\mu = 0.14$ | RCut | **29.75** | − 3.53% | − 15.3% | **29.75** | − 20.3% | NCut | **2.92** | **2.92** | **2.92** | **2.92** | − 36.2% |
| | ACC | **1.0** | − 8.2% | − 10.4% | **1.0** | − 7.2% | ACC | **1.0** | **1.0** | **1.0** | **1.0** | − 8.5% |
| | NMI | **1.0** | − 2.4% | − 4.6% | **1.0** | − 4.7% | NMI | **1.0** | **1.0** | **1.0** | **1.0** | − 5.4% |
| $\mu = 0.16$ | RCut | **31.37** | − 4.6% | **31.37** | **31.37** | − 3.0% | NCut | **3.17** | **3.17** | **3.17** | **3.17** | − 13.5% |
| | ACC | **1.0** | − 2.8% | **1.0** | **1.0** | − 3.1% | ACC | **1.0** | **1.0** | **1.0** | **1.0** | − 2.6% |
| | NMI | **1.0** | − 1.7% | **1.0** | **1.0** | − 1.6% | NMI | **1.0** | **1.0** | **1.0** | **1.0** | − 1.8% |
| $\mu = 0.18$ | RCut | **30.56** | − 0.15% | − 1.82% | **30.56** | **30.56** | NCut | **3.21** | − 0.1% | **3.21** | **3.21** | − 11.8% |
| | ACC | **1.0** | − 0.1% | − 0.7% | **1.0** | **1.0** | ACC | **1.0** | − 0.1% | **1.0** | **1.0** | − 2.5% |
| | NMI | **1.0** | − 0.2% | − 1.1% | **1.0** | **1.0** | NMI | **1.0** | − 0.19% | **1.0** | **1.0** | − 1.6% |
| $\mu = 0.20$ | RCut | **34.45** | − 4.1% | − 2.77% | **34.45** | − 19.0% | NCut | **3.65** | **3.65** | − 1.53% | **3.65** | − 11.1% |
| | ACC | − 0.1% | − 3.2% | − 12.0% | **1.0** | − 24.1% | ACC | **1.0** | − 0.1% | − 0.5% | **1.0** | − 2.4% |
| | NMI | **1.0** | − 2.1% | − 5.2% | **1.0** | − 17.0% | NMI | **1.0** | − 0.19% | − 0.9% | **1.0** | − 2.4% |
| $\mu = 0.22$ | RCut | **39.01** | − 5.0% | − 4.0% | − 0.34% | − 1.00% | NCut | **3.96** | **3.96** | **3.96** | **3.96** | **3.96** |
| | ACC | **1.0** | − 15.0% | − 11.0% | **1.0** | − 7.2% | ACC | **1.0** | − 0.3% | **1.0** | **1.0** | **1.0** |
| | NMI | **1.0** | − 7.2% | − 6.7% | − 0.5% | − 3.3% | NMI | **1.0** | − 0.2% | − 0.9% | **1.0** | **1.0** |
| $\mu = 0.24$ | RCut | **49.76** | − 7.3% | − 7.2% | − 1.4% | − 2.3% | NCut | **5.02** | **5.02** | **5.02** | **5.02** | − 8.8% |
| | ACC | − 13.8% | **0.88** | − 12.8% | − 10.1% | − 39.0% | ACC | **1.0** | − 0.3% | **1.0** | **1.0** | − 3.3% |
| | NMI | − 0.46% | **0.94** | − 7.3% | − 3.8% | − 16.7% | NMI | **1.0** | − 0.19% | − 0.9% | − 0.2% | − 1.9% |
| $\mu = 0.26$ | RCut | **41.95** | − 4.5% | − 1.2% | − 1.9% | − 11.4% | NCut | **4.4** | − 0.14% | − 41.0% | **4.4** | **4.4** |
| | ACC | − 1.9% | − 5.2% | − 13.7% | **0.87** | − 48.0% | ACC | **1.0** | − 0.1% | − 9.2% | **1.0** | **1.0** |
| | NMI | − 1.0% | − 7.3% | − 8.7% | **0.94** | − 32.5% | NMI | **1.0** | − 0.17% | − 14.7% | − 0.18% | **1.0** |

**Table 2** (continued)

| Case | Measure | pGrass | Spec | pSpec | kCuts | Luo | Measure | pGrass | Spec | pSpec | kCuts | Graclus |
|---|---|---|---|---|---|---|---|---|---|---|---|---|
| $\mu = 0.28$ | RCut | **53.38** | − 8.4% | − 5.2% | − 4.4% | − 1.0% | NCut | **5.59** | − 0.1% | − 0.4% | 5.59 | − 6.6% |
| | ACC | **0.83** | − 44.4% | − 19.6% | 0.83 | − 152.7% | ACC | **1.0** | − 0.2% | − 0.4% | **1.0** | − 3.1% |
| | NMI | − 0.3% | − 23.3% | − 6.3% | **0.894** | − 78.3% | NMI | **1.0** | − 0.5% | − 0.7% | − 0.18% | − 1.8% |
| $\mu = 0.30$ | RCut | − 1.74% | − 10.5% | − 10.9% | − 4.9% | **56.0** | NCut | **6.00** | − 0.14% | − 1.96% | **6.00** | **6.00** |
| | ACC | **0.77** | − 26.4% | − 68.2% | 0.77 | − 67.1% | ACC | **1.0** | − 0.3% | − 1.9% | **1.0** | **1.0** |
| | NMI | **0.866** | − 18.3% | − 37.8% | − 0.2% | − 34.2% | NMI | **1.0** | − 0.52% | − 3.0% | − 0.74% | **1.0** |
| $\mu = 0.32$ | RCut | **43.73** | − 13.5% | − 9.4% | − 4.9% | − 0.35% | NCut | **5.40** | − 0.48% | − 1.0% | − 0.04% | − 11.8% |
| | ACC | **0.51** | − 3.7% | − 17.5% | 0.51 | − 126.1% | ACC | **0.998** | − 0.8% | − 2.0% | − 0.2% | − 6.9% |
| | NMI | − 1.6% | − 6.4% | **0.67** | − 0.3% | − 95.0% | NMI | **0.996** | − 1.5% | − 3.5% | − 0.34% | − 4.5% |
| $\mu = 0.34$ | RCut | − 1.47% | − 9.7% | − 14.0% | − 6.5% | **46.24** | NCut | − 0.1% | − 0.6% | − 1.4% | − 0.1% | **5.74** |
| | ACC | − 7.9% | **0.53** | − 57.4% | − 94.9% | − 247.7% | ACC | − 0.1% | − 0.8% | − 1.7% | **0.995** | **0.995** |
| | NMI | **0.62** | − 0.7% | − 4.7% | − 35.6% | − 213.9% | NMI | − 0.19% | − 1.6% | − 2.9% | − 0.01% | **0.990** |
| $\mu = 0.36$ | RCut | − 1.3% | − 6.3% | − 19.6% | − 4.5% | **56.11** | NCut | − 0.04% | − 1.00% | − 5.9% | **7.216** | − 0.02% |
| | ACC | **0.38** | − 7.8% | − 15.4% | **0.38** | − 51.2% | ACC | − 0.6% | − 2.3% | − 7.2% | **0.998** | − 0.5% |
| | NMI | **0.51** | − 2.4% | − 8.0% | − 1.4% | − 42.1% | NMI | − 1.1% | − 3.7% | − 11.0% | **0.996** | − 0.74% |
| $\mu = 0.38$ | RCut | − 0.08% | − 10.3% | − 14.7% | − 0.2% | **56.25** | NCut | **7.18** | − 1.0% | − 6.8% | − 0.08% | − 4.5% |
| | ACC | − 0.9% | **0.34** | − 54.4% | − 43.8% | − 117.5% | ACC | **0.992** | − 2.0% | − 9.0% | − 0.4% | − 4.4% |
| | NMI | **0.51** | − 12.6% | − 36.3% | − 61.0% | − 156.0% | NMI | **0.996** | − 3.8% | − 14.2% | − 0.76% | − 3.6% |
| $\mu = 0.40$ | RCut | **53.45** | − 10.9% | − 18.9% | − 1.9% | − 1.4505 | NCut | − 0.2% | − 0.9% | − 8.1% | **7.99** | − 3.5% |
| | ACC | **0.25** | − 12.8% | **0.25** | − 28.1% | − 39.0% | ACC | **0.998** | − 2.1516% | − 12.2% | − 0.8% | − 3.9% |
| | NMI | **0.43** | − 64.2% | − 58.6% | − 69.2% | − 102.0% | NMI | − 1.55% | − 4.1% | − 17.1% | **0.994%** | − 3.7% |

The numerical value of the best solutions for each metric is presented in bold. The percentage signs indicate how much inferior a solution is to the best solution

**Table 3** Clustering results for the Omniglot database

| Measure | Case | pGr-A | pGr-B | Spec | pSpec | kCuts | Graclus | Case | pGr-A | pGr-B | Spec | pSpec | kCuts | Graclus |
|---|---|---|---|---|---|---|---|---|---|---|---|---|---|---|
| NCut | AngloSaxon | 13.63 | −2.57% | −3.54% | −8.95% | −2.89% | −0.63% | Ojibwe | −1.80% | −3.50% | −6.10% | −12.62% | 4.85 | −3.68% |
| ACC | | −5.04% | 0.358 | −2.46% | −30.83% | −14.28% | −11.23% | | −2.65% | 0.41 | −0.88% | −0.88% | −8.43% | −7.42% |
| NMI | | −2.14% | 0.477 | −1.06% | −9.99% | −5.16% | −5.68% | | −0.59% | 0.46 | −1.61% | −1.07% | −2.51% | −3.90% |
| NCut | Arcadian | 15.42 | −1.86% | −2.12% | −6.23% | −1.05% | −0.22% | Sanskrit | −0.50% | −2.42% | −2.19% | −5.43% | −3.30% | 0.319% |
| ACC | | −2.85% | 0.29 | −0.10% | −8.04% | −2.85% | −5.02% | | −1.46% | 0.166 | 0.166 | −5.31% | −2.96% | −6.86% |
| NMI | | −4.89% | 0.384 | −2.32% | −16.53% | −7.29% | −5.20% | | −1.39% | 0.33 | −1.33% | −9.70% | −4.27% | −4.66% |
| NCut | Armenian | 23.02 | −2.64% | −2.83% | −4.20% | −0.96% | −1.12% | Syriac | 12.59 | −3.00% | −4.11% | −3.71% | −1.70% | −3.59% |
| ACC | | 0.24 | −2.65% | −4.88% | −10.87% | −3.77% | −3.18% | | −8.01% | 0.263 | −4.28% | −17.46% | −8.01% | −7.04% |
| NMI | | −1.08% | 0.40 | −1.08% | −8.18% | −4.16% | −2.43% | | −8.10% | 0.338 | −2.57% | −18.41% | −10.21% | −8.80% |
| NCut | Balinese | 14.31 | −3.42% | −4.92% | −5.79% | −2.14% | −2.59% | Tifinagh | 28.84 | −5.21% | −7.44% | −6.45% | −3.83% | −3.05% |
| ACC | | −4.16% | −4.16% | −6.85% | −8.68% | 0.26 | −2.44% | | −1.68% | 0.279 | −0.98% | −13.69% | −9.24% | −2.01% |
| NMI | | −4.02% | −2.84% | −7.09% | −8.56% | −3.2% | 0.33 | | −4.97% | 0.498 | −0.97% | −9.63% | −8.74% | −4.42% |
| NCut | Bengali | 26.31 | −1.88% | −2.34% | −6.52% | −1.20% | −0.75% | Angelic | 8.90 | −4.99% | −6.97% | −6.89% | −1.61% | −5.15% |
| ACC | | −8.57% | 0.223 | −4.16% | −19.25% | −2.06% | −10.29% | | −11.90% | −3.30% | 0.47 | −20.51% | −12.57% | −14.63% |
| NMI | | −4.41% | −0.75% | −1.36% | −9.46% | 0.40 | −2.73% | | −5.86% | 0.538 | −0.73% | −18.12% | −9.14% | −6.55% |
| NCut | Blackfoot | 4.39 | −2.87% | −7.29% | −5.78% | −0.81% | −10.92% | Atemayar | −1.09% | −3.08% | −5.85% | −7.18% | −2.44% | 13.26 |
| ACC | | −7.93% | −4.82% | −6.86% | −17.22% | −4.82% | 0.389 | | −7.30% | 0.26 | −2.44% | −15.56% | −9.01% | −18.61% |
| NMI | | −8.24% | −4.12% | −8.92% | −9.02% | −0.27% | 0.452 | | −4.62% | 0.351 | −0.54% | −12.14% | −6.36% | −8.67% |
| NCut | Braille | 13.76 | −4.83% | −6.82% | −4.90% | −1.92% | −1.99% | Atlantean | 16.49 | −4.76% | −5.54% | −4.86% | −1.16% | −0.39% |
| ACC | | −16.69% | 0.298 | −2.12% | −27.24% | −13.28% | −17.61% | | −12.05% | 0.34 | 0.34 | −18.81% | −14.92% | −5.35% |
| NMI | | −11.61% | 0.387 | −1.76% | −18.21% | −12.71% | −11.90% | | −8.53% | −0.79% | −3.12% | −16.91% | −11.42% | 0.422 |
| NCut | Burmese | 19.40 | −3.76% | −3.89% | −5.73% | −1.48% | −1.30% | Aurek-Besh | 11.66 | −4.92% | −6.80% | −5.25% | −1.82% | −2.72% |
| ACC | | −5.39% | 0.23 | −1.32% | −15.45% | 0.23 | −3.96% | | −3.21% | 0.37 | −3.21% | −21.53% | −1.04% | −8.46% |
| NMI | | −3.50% | −0.43% | −0.89% | −9.92% | −4.40% | 0.372 | | −2.34% | 0.47 | −1.67% | −9.71% | −2.05% | −5.24% |
| NCut | Cyrillic | 16.79 | −3.28% | −4.25% | −8.08% | −0.80% | −2.69% | Avesta | 15.25 | −3.67% | −4.53% | −6.37% | −2.40% | −1.86% |
| ACC | | −0.45% | −0.18% | 0.334 | −19.44% | −0.90% | −9.38% | | −6.11% | 0.27 | −0.72% | −21.94% | −1.44% | −7.74% |
| NMI | | −2.91% | 0.467 | −0.65% | −15.15% | −2.64% | −4.19% | | −6.36% | 0.366 | −2.00% | −20.24% | −8.14% | −5.83% |

**Table 3** (continued)

| Measure | Case | pGr-A | pGr-B | Spec | pSpec | kCuts | Graclus | Case | pGr-A | pGr-B | Spec | pSpec | kCuts | Graclus |
|---|---|---|---|---|---|---|---|---|---|---|---|---|---|---|
| NCut | Aramaic | **10.37** | −4.34% | −4.35% | −8.21% | −2.58% | −3.40% | Ge-ez | −0.48% | −3.36% | −4.02% | −3.57% | −1.20% | **15.11%** |
| ACC | | −1.32% | **0.352** | −4.73% | −13.98% | −2.65% | −3.34% | | −10.00% | **0.275** | **0.275** | −16.28% | −17.22% | −15.30% |
| NMI | | −0.28% | **0.427** | −0.99% | −16.10% | −0.12% | −2.64% | | −4.42% | −5.41% | **0.355** | −6.29% | −6.55% | −0.82% |
| NCut | Futurama | **15.97** | −5.47% | −6.89% | −3.90% | −0.83% | −1.46% | Glagolitic | **27.84** | −3.96% | −5.82% | −5.11% | −1.52% | −0.03% |
| ACC | | −6.53% | **0.344** | −1.12% | −13.30% | −14.01% | −15.46% | | −8.02% | **0.3** | −1.12% | −23.97% | −5.10% | −10.25% |
| NMI | | −9.48% | **0.446** | −0.59% | −8.21% | −15.66% | −9.35% | | −5.13% | **0.45** | −1.21% | −19.61% | −4.14% | −2.25% |
| NCut | Georgian | **18.70** | −1.63% | −4.70% | −5.19% | −2.98% | −2.38% | Gurmukhi | **24.96** | −2.17% | −3.62% | −5.87% | −1.25% | −2.79% |
| ACC | | −2.89% | **0.313** | −1.62% | −11.08% | −8.21% | −3.29% | | −1.16% | −5.52% | −8.15% | −5.52% | −5.52% | **0.19** |
| NMI | | −1.92% | **0.488** | −1.66% | −4.82% | −5.23% | −1.92% | | −2.27% | −1.01% | −1.61% | −3.98% | −1.24% | **0.36** |
| NCut | Grantha | **22.37** | −2.06% | −2.52% | −5.40% | −3.20% | −0.89% | Kannada | −0.26% | −3.01% | −3.68% | −5.51% | −1.21% | **24.32** |
| ACC | | −0.35% | −0.50% | **0.341** | −9.72% | −3.54% | −6.91% | | −2.96% | **0.21** | −5.47% | −4.17% | −9.44% | −2.96% |
| NMI | | −0.14% | −2.05% | **0.498** | −4.71% | −3.99% | −2.62% | | −2.35% | **0.37** | −0.75% | −4.23% | −5.61% | −0.70% |
| NCut | Greek | **13.19** | −3.60% | −6.96% | −6.47% | −1.15% | −1.12% | Keble | **8.91** | −2.60% | −4.86% | −6.40% | −0.26% | −3.59% |
| ACC | | **0.329** | −2.75% | −5.34% | −22.52% | −11.29% | −8.22% | | −2.57% | **0.30** | 0.00% | −10.47% | −3.26% | −2.57% |
| NMI | | **0.410** | −2.91% | −2.04% | −13.63% | −6.00% | −2.60% | | −3.40% | −0.30% | −1.36% | −10.41% | −4.77% | **0.402** |
| NCut | Gujarati | **29.86** | −2.62% | −3.44% | −4.83% | −0.86% | −0.33% | Malayalam | **29.38** | −1.71% | −3.49% | −4.66% | −0.95% | −0.39% |
| ACC | | −1.54% | **0.21** | −3.57% | −14.70% | −4.14% | −10.33% | | −5.58% | **0.22** | −1.98% | −10.65% | −0.50% | −6.14% |
| NMI | | −3.08% | **0.41** | −1.13% | −8.74% | −4.20% | −3.69% | | −7.34% | **0.41** | −1.72% | −13.90% | −6.25% | −9.88% |
| NCut | Hebrew | **11.73** | −4.09% | −4.29% | −4.49% | −1.67% | −0.31% | Manipuri | −0.16% | −2.07% | −4.92% | −4.88% | −1.08% | **24.08** |
| ACC | | −9.37% | **0.248** | −5.16% | −16.39% | −3.20% | −4.19% | | −7.23% | **0.22** | −1.69% | −16.31% | −2.30% | −7.23% |
| NMI | | −4.18% | **0.333** | −0.33% | −13.11% | −4.08% | −6.20% | | −2.02% | **0.369** | −0.22% | −10.42% | −0.55% | −5.40% |
| NCut | Hiragana | −0.70% | −3.96% | −4.21% | −7.52% | −2.01% | **33.68** | Mongolian | **17.49** | −2.31% | −4.18% | −5.07% | −1.33% | −2.51% |
| ACC | | −6.25% | −8.30% | −2.36% | −13.62% | **0.268** | −9.58% | | −7.41% | **0.27** | −1.91% | −20.30% | −11.92% | −7.41% |
| NMI | | −6.40% | **0.452** | −1.16% | −9.18% | −3.93% | −5.90% | | −4.27% | **0.39** | −2.39% | −12.58% | −8.48% | −5.03% |

**Table 3** (continued)

| Measure | Case | pGr-A | pGr-B | Spec | pSpec | kCuts | Graclus | Case | pGr-A | pGr-B | Spec | pSpec | kCuts | Graclus |
|---|---|---|---|---|---|---|---|---|---|---|---|---|---|---|
| NCut | Inuktitut | −4.73% | **4.55** | −10.97% | −9.34% | −5.28% | −7.74% | Slavonic | **29.29** | −3.53% | −4.53% | −4.10% | −1.22% | −1.14% |
| ACC | | −6.44% | **0.515** | −1.86% | −9.26% | −0.60% | −22.21% | | −5.94% | **0.24** | −2.48% | −20.34% | −5.94% | −9.03% |
| NMI | | −2.57% | **0.582** | −1.25% | −7.59% | −2.37% | −6.95% | | −6.85% | **0.41** | −1.72% | −17.94% | −6.76% | −5.30% |
| NCut | Katakana | **26.18** | −3.49% | −4.70% | −6.39% | −1.32% | −1.12% | Oriya | **28.63** | −2.68% | −3.06% | −4.56% | −0.79% | −0.55% |
| ACC | | −6.43% | **0.246** | −4.49% | −7.40% | −10.47% | −9.45% | | −6.89% | **0.203** | −1.09% | −5.06% | −3.88% | −10.01% |
| NMI | | −5.29% | **0.435** | −1.63% | −10.72% | −5.32% | −3.54% | | −4.33% | **0.37** | −0.46% | −5.36% | −2.95% | −2.61% |
| NCut | Korean | −0.04% | −2.26% | −3.20% | −7.17% | −1.76% | **22.47** | Sylhetti | **14.43** | −2.10% | −4.00% | −5.58% | −0.72% | −1.05% |
| ACC | | **0.23** | −3.31% | −5.65% | −16.90% | −2.19% | −1.08% | | −3.68% | −6.67% | −0.91% | −2.77% | −14.29% | **0.2%** |
| NMI | | −1.51% | **0.40** | −3.01% | −11.13% | −3.99% | −1.97% | | −5.43% | **0.293** | −1.52% | −8.19% | −9.36% | −0.93% |
| NCut | Magi | **11.91** | −3.66% | −4.79% | −3.54% | −0.34% | −1.32% | Syriac-Serto | **12.02** | −5.93% | −6.23% | −4.66% | −3.97% | −3.99% |
| ACC | | −9.09% | −0.76% | **0.33** | −13.79% | −3.94% | −13.79% | | −5.97% | **0.31** | **0.308** | −8.39% | −14.50% | −3.66% |
| NMI | | −3.32% | −1.00% | −2.36% | −14.31% | **0.373** | −2.98% | | −2.58% | −2.99% | **0.41** | −14.43% | −16.87% | −5.46% |
| NCut | Latin | **12.76** | −4.37% | −4.42% | −5.08% | −2.15% | −1.46% | Tengwar | **13.7** | −2.43% | −4.42% | −7.76% | −1.25% | −2.28% |
| ACC | | −6.35% | **0.296** | −1.44% | −7.82% | −10.92% | −7.82% | | −7.91% | **0.30** | −4.17% | −7.14% | −4.17% | −7.91% |
| NMI | | −4.58% | −2.18% | −1.26% | −9.13% | −7.29% | **0.426** | | −6.39% | **0.37** | −0.73% | −5.70% | −2.33% | −0.05% |
| NCut | Malay | **21.85** | −3.43% | −4.72% | −7.05% | −1.94% | −2.88% | Tibetan | **25.38** | −2.76% | −2.76% | −5.59% | −1.06% | −0.30% |
| ACC | | **0.213** | **0.213** | −10.27% | −10.27% | **0.213** | −2.99% | | **0.24** | −3.11% | −3.65% | −21.49% | −5.32% | −6.46% |
| NMI | | −4.27% | **0.395** | −0.89% | −8.39% | −2.86% | −1.70% | | −0.67% | **0.41** | −0.77% | −14.48% | −4.88% | −6.54% |
| NCut | Mkhedruli | **24.10** | −3.21% | −3.73% | −4.96% | −1.38% | −2.40% | ULOG | **14.29** | −3.49% | −4.29% | −2.41% | −1.37% | −2.71% |
| ACC | | −13.96% | **0.25** | −0.60% | −16.49% | −16.49% | −6.30% | | −3.87% | **0.26** | −0.74% | −16.50% | −3.87% | −11.66% |
| NMI | | −8.05% | **0.42** | −1.70% | −15.27% | −9.57% | −5.59% | | −5.56% | **0.36** | −1.06% | −13.17% | −5.13% | −10.47% |
| NCut | N−Ko | **14.91** | −3.26% | −4.41% | −6.80% | −1.29% | −3.09% | Tagalog | −0.10% | −2.98% | −5.69% | −7.39% | −1.11% | **8.71** |
| ACC | | −2.45% | **0.318** | −0.47% | −9.38% | −1.95% | −7.14% | | **0.37** | −7.57% | −2.42% | −17.44% | −8.47% | −7.57% |
| NMI | | −1.24% | **0.464** | −0.43% | −8.78% | −2.67% | −1.51% | | −1.37% | **0.41** | −0.39% | −6.96% | −3.17% | −0.69% |

The numerical value of the best solutions for each metric is presented in bold. The percentage signs indicate how much inferior a solution is to the best solution. pGr-A and pGr-B refer to the algorithmic variants pGrass-kmeans and pGrass-disc respectively

**Author contributions** All authors contributed to the conception and the design of the method. Material preparation, data collection and analysis were performed by Dimosthenis Pasadakis and Christie Louis Alappat. A critical revision of the article with new interpretations of the results was performed by Olaf Schenk and Gerhard Wellein. All authors reviewed the results and approved the final version of the manuscript.

**Funding** Open access funding provided by Università della Svizzera italiana. We would like to acknowledge the financial support from the Swiss National Science Foundation (SNSF) under the project 182673 entitled *Balanced Graph Partition Refinement Using the Graph p-Laplacian*.

**Availability of data and material** The data used in the experiments is available upon request to the first author.

**Code availability** The code written to obtain the numerical results is available upon request to the first author.

## Declarations